\documentclass[final]{cvpr}

\usepackage{times}
\usepackage{epsfig}
\usepackage{graphicx}
\usepackage{amsmath}
\usepackage{amssymb}
\usepackage{pifont}
\usepackage{multirow}
\usepackage{enumitem}
\usepackage[dvipsnames,table,xcdraw]{xcolor}
\usepackage{subcaption}
\usepackage{makecell}
\usepackage{tablefootnote}
\usepackage{textpos}

\newcommand{\YL}{\textcolor{black}}

\newcommand{\cc}{\textcolor{black}}

\newcommand{\cmark}{\ding{51}}%
\newcommand{\xmark}{\ding{53}}%

\usepackage[pagebackref=true,breaklinks=true,colorlinks,bookmarks=false]{hyperref}

\def\cvprPaperID{1397} 
\def\confYear{CVPR 2021}
\begin{document}

\title{Ego-Exo: Transferring Visual Representations\\from Third-person to First-person Videos}


\author{
Yanghao Li$^{1}$~~~
Tushar Nagarajan$^{1,2}$~~~
Bo Xiong$^{1}$~~~
Kristen Grauman$^{1,2}$ \\
$^{1}$ Facebook AI Research
$^{2}$ UT Austin~~~\\
\tt\small lyttonhao@fb.com, tushar@cs.utexas.edu, bxiong@fb.com, grauman@fb.com
}

\maketitle
\thispagestyle{empty}

\begin{abstract}
We introduce an approach for pre-training \emph{egocentric} video models using large-scale \emph{third-person} video datasets. Learning from purely egocentric data is limited by low dataset scale and diversity, while using purely exocentric (third-person) data introduces a large domain mismatch. 
Our idea is to discover latent signals in third-person video that are predictive of key egocentric-specific properties. Incorporating these signals as knowledge distillation losses during pre-training results in models that benefit from both the scale and diversity of third-person video data, as well as representations that capture salient egocentric properties. 
Our experiments show that our ``Ego-Exo'' framework can be seamlessly integrated into standard video models; it outperforms all baselines when fine-tuned for egocentric activity recognition, achieving state-of-the-art results on Charades-Ego and EPIC-Kitchens-100.


\end{abstract}

\section{Introduction}

\begin{textblock*}{\textwidth}(0cm,-16cm)
\centering
In Proceedings of the IEEE Conference on Computer Vision and Pattern Recognition (CVPR), 2021. 
\end{textblock*}


Egocentric video captured by wearable cameras offers a unique perspective into human behavior.  It is the subject of a recent surge in research interest in first-person activity recognition~\cite{epic-fusion,zhou2015temporal}, anticipation~\cite{furnari2020rolling,abu2018will}, and video summarization~\cite{lee2015predicting,yonetani2016visual,del2016summarization}
with many valuable future applications in augmented reality and robotics.
Compared to third-person videos, 
egocentric videos show the world through a distinct viewpoint, encode characteristic egocentric motion patterns due to body and head movements, and have a unique focus on hands, objects, and faces, driven by the camera wearer's attention and interaction with their surroundings.


However, these unique properties also present a fundamental challenge for video understanding.
On the one hand, learning models purely from egocentric data are limited by dataset scale. Current egocentric video datasets are small (e.g., 90k clips in EPIC-Kitchens-100~\cite{epic-100} vs.~650k in Kinetics-700~\cite{kinetics}) and lack diversity (e.g., videos only in kitchen scenes). 
On the other hand, a purely exocentric approach that uses more readily available third-person 
video---the status-quo for pre-training video models~\cite{slowfast,tsn,zhou2018temporal,nonlocal}---ignores the unique properties 
of egocentric video and faces a major domain mismatch.
Prior work has shown that this latter strategy, though popular, is insufficient: pre-training egocentric action recognition models with third-person data alone produces significantly worse results than pre-training with first-person data~\cite{charades-ego}.
In an attempt to bridge the domain gap, prior work explores traditional embedding learning~\cite{sigurdsson2018actor,yu2019see} or domain adaptation approaches~\cite{choi2020unsupervised}, but they require \emph{paired} egocentric and third-person videos that are either concurrently recorded or annotated for the same set of activities, which are difficult to collect and hence severely limit their scope.

\begin{figure}[t!]
\centering
\includegraphics[width=\columnwidth]{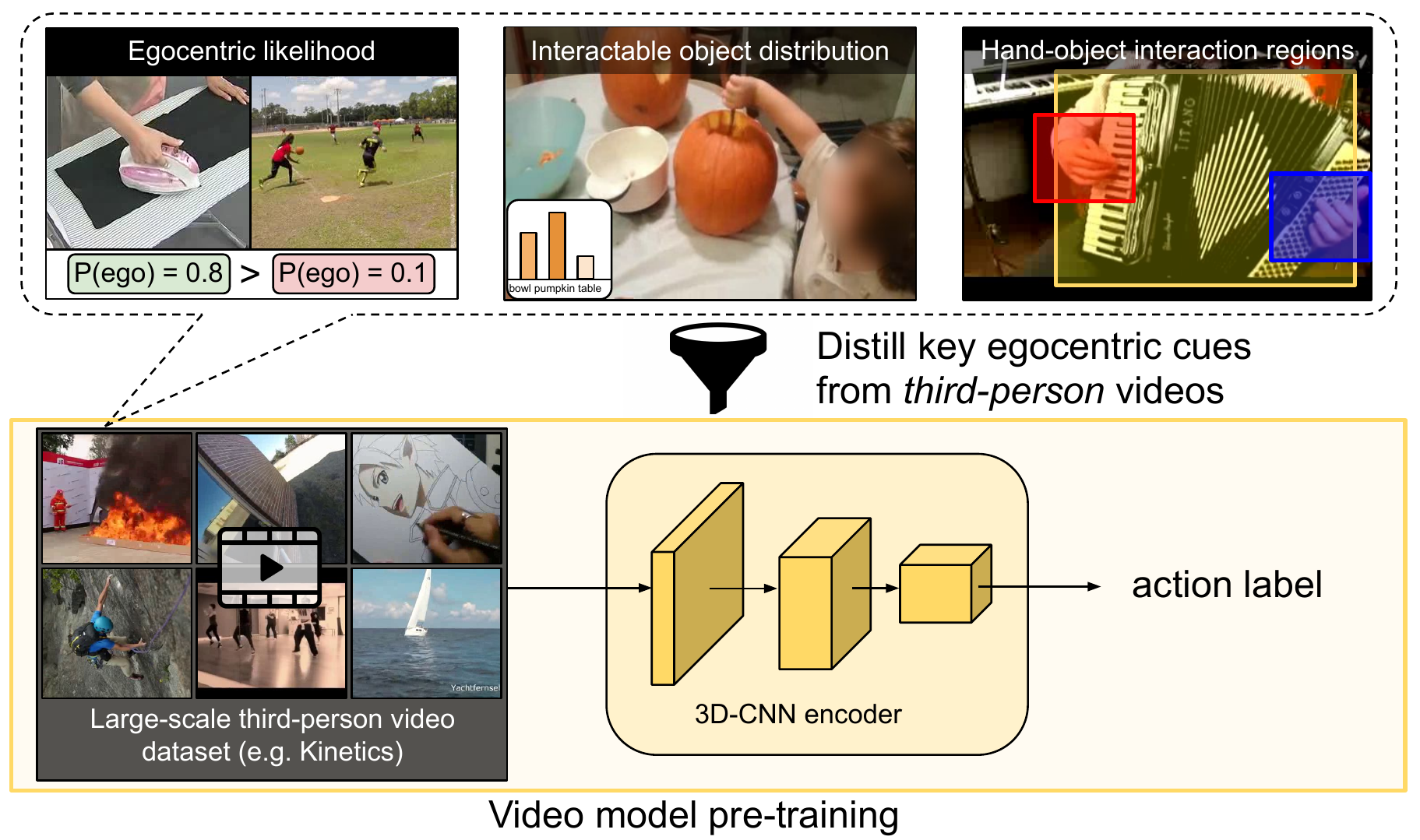}
\caption{\textbf{Main idea.} We extract key egocentric signals from large-scale third-person data and distill them into the video backbone during pre-training to guide feature learning for egocentric video tasks with wearable camera video.}
\vspace{-4.5mm}
\label{fig:main-idea}
\end{figure}



Despite their differences, we hypothesize that the exocentric view of activity should in fact inform the egocentric view.   First, humans are able to watch videos of other people performing activities and map actions into their own (egocentric) perspective; babies in part learn new skills in just this manner~\cite{meltzoff1988imitation,premack1978does}.
Second, exocentric video is not devoid of person-centered cues.  For example, a close-up instructional video captured from the third-person view may nonetheless highlight substantial hand-object interactions; or video captured with a hand-held phone may follow an event (e.g., a parade) as it unfolds with attentional cues related to a head-mounted camera.  



Building on this premise, in this work we ask: \emph{``How can we best utilize current video datasets to pre-train egocentric video models?''}  Our key idea is to discover latent signals in third-person video that approximate egocentric-specific properties.  To that end, we introduce a feature learning approach in which ego-video features are guided by both exo-video activity labels and (unlabeled) ego-video cues, to better align traditional third-person video pre-training with downstream egocentric video tasks.
Specifically, we introduce a series of ego-inspired tasks that require the video model to be predictive of manipulated objects, spatiotemporal hand-object interaction regions, and general egocentricity.  Then we incorporate these tasks into training as knowledge-distillation losses to supplement  
an action classification pre-training objective on third-person video. See Fig.~\ref{fig:main-idea}.


By design, our video models can continue to enjoy large amounts of labeled third-person training data, while simultaneously embedding egocentric signals into the learned features, making them a suitable drop-in replacement for traditional video encoders for egocentric video tasks.
Finally, our approach does not assume any paired or activity-labeled egocentric videos during pre-training; the egocentric signals are directly inferred from third-person video.


Our experiments on three challenging egocentric video datasets show that our ``Ego-Exo'' framework learns strong egocentric feature representations 
from third-person video. 
On Charades-Ego~\cite{charades-ego}, our model improves over models pre-trained on Kinetics---the standard 
pre-training and fine-tuning paradigm---by +3.26 mAP, and outperforms methods that specifically aim to bridge the domain gap between viewpoints. 
Finally, our pre-trained model achieves state-of-the-art results on  EPIC-Kitchens-100~\cite{epic-100}, the largest available first-person dataset. 




\begin{section}{Related Work}


\paragraph{Egocentric video understanding}
The unique viewpoint in egocentric video presents interesting research challenges including action recognition and anticipation~\cite{zhou2015temporal,furnari2020rolling,rhinehart2017first,abu2018will}, daily life summary generation~\cite{lee2015predicting,yonetani2016visual}, inferring body pose~\cite{jiang2017seeing,ng2020you2me}, and estimating gaze~\cite{li2013learning,huang2018predicting}. Several egocentric video datasets have been created to support these challenges~\cite{epic,li2018eye,pirsiavash2012detecting,charades-ego}. Model architectures proposed for these tasks include multi-stream networks~\cite{ma2016going,li2018eye,epic-fusion,wang2020makes}, recurrent networks~\cite{furnari2019would,furnari2020rolling,sudhakaran2019lsta}, 3D conv nets~\cite{pirri2019anticipation,lu2019learning} and spatially grounded topological graph models~\cite{ego-topo}.  

These architectures vary significantly, but all use video encoders that are similarly pre-trained with third-person video datasets, despite being applied to egocentric video tasks. In contrast, we introduce key egocentric losses during \emph{exocentric video} pre-training that bridge the domain gap when applied to downstream egocentric video tasks. 

\vspace{-3mm}
\paragraph{Joint first/third person video understanding} 

Several strategies have been proposed to address the domain gap between first and third person video. Prior work learns \emph{viewpoint-invariant} representations using embedding learning methods, and applies them to action recognition~\cite{soran2014action,sigurdsson2018actor,ardeshir2018exocentric}, video summarization~\cite{ho2018summarizing}, image retrieval~\cite{fan2017identifying}, person segmentation~\cite{xu2018joint}, and attention-driven gaze prediction~\cite{yu2019see}. 
Image generation methods~\cite{elfeki2018third,regmi2018cross,regmi2019bridging,liu2020exocentric} use generative adversarial frameworks to synthesize one viewpoint from the other.  Viewpoint invariance has also been treated as a \emph{domain adaptation} task in prior work, adapting third-person video models for overhead drone-footage~\cite{choi2020unsupervised}.
Other methods use egocentric video as a modality to supplement top-view footage to improve identification and tracking models~\cite{ardeshir2016ego2top, ardeshir2018integrating,ardeshir2016egoreid,yang2018ego}. 

The above methods require \emph{paired} datasets that are either simultaneously recorded or that share the same labels for instances across viewpoints. In contrast, our method leverages only third-person video datasets, but is augmented with pseudo-labels (derived from first-person models) to learn egocentric video representations, thus circumventing both the need for first-person video during pre-training and the need for paired labeled data.




\vspace{-3mm}
\paragraph{Knowledge distillation for video} 







In knowledge distillation (KD), one network is trained to reproduce the outputs of another~\cite{hinton2015distilling}.  Distillation 
serves to compress 
models~\cite{hinton2015distilling,mullapudi2019online,chen2017learning} or incorporate privileged information from alternate tasks~\cite{lopez2015unifying}. In videos, KD can incorporate information from alternate modalities like audio~\cite{gao2020listen,aytar2016soundnet}, depth and flow~\cite{gupta2016cross,stroud2020d3d}, or a combination
~\cite{luo2018graph}, 
and 
object-level information~\cite{pan2020spatio}.
In the context of self-supervised learning, prior work assigns weak image labels to video instances as supervision for video-level models~\cite{girdhar2019distinit}. In contrast, we use inferred weak labels 
that are relevant to the egocentric domain, 
and we use them alongside third-person video labels, rather than in place of them during pre-training.





\vspace{-3mm}
\paragraph{Egocentric cues in video understanding models}
Egocentric video offers several unique cues that have been leveraged to improve video understanding models~\cite{li2015delving,ma2016going}. These include attention mechanisms from gaze and motor attention~\cite{mathe2012dynamic,li2018eye,liu2019forecasting}, active object detection~\cite{antonino-next-active,baradel2018object,dessalene2020egocentric,wang2020symbiotic}, and hands in contact~\cite{tekin2019h,shan2020understanding,kapidis2019egocentric}.
We are also interested in such important egocentric cues, but unlike prior work we do not train models on labeled egocentric video datasets 
to detect them. Instead, we embed labels predicted for these cues as auxiliary losses in \emph{third-person video} models to steer feature learning towards egocentric relevant features. 
Unlike any of these prior methods, our goal is to leverage third-person video to pre-train first-person video models.

\end{section}
\begin{figure*}[t!]
\centering
\includegraphics[width=1.95\columnwidth]{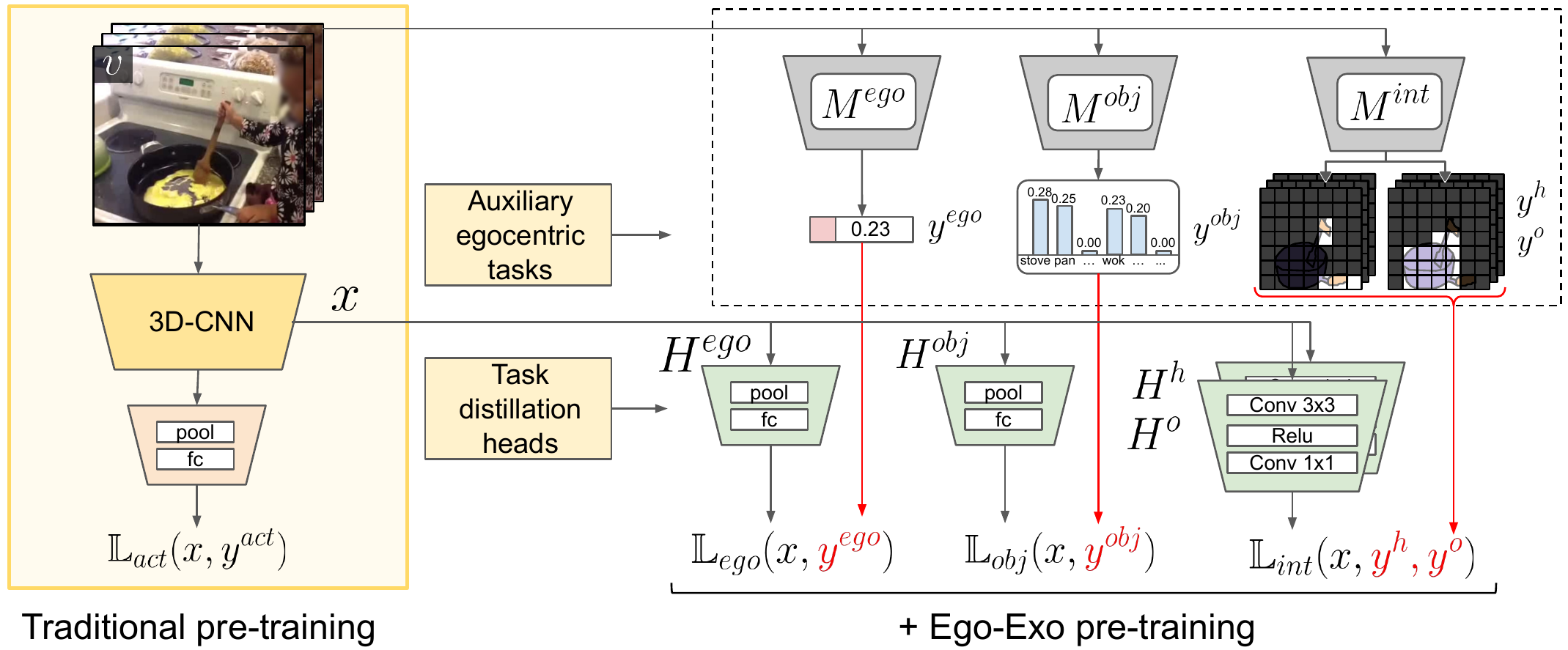}
\vspace{-1.5mm}
\caption{\textbf{Ego-Exo framework}. To enhance traditional pre-training (left panel), we generate soft-labels for \emph{third-person videos} from a set of pre-trained egocentric models (top-right) that capture a variety of key egocentric signals (Sec~\ref{sec:tasks}), and we train distillation modules to approximate the responses of these models (bottom-right). 
Once pre-trained, the video backbone can be directly fine-tuned for a downstream egocentric task.
}
\vspace{-5mm}
\label{fig:framework}
\end{figure*}

\section{Ego-Exo Approach}


Our goal is to learn egocentric video representations from third-person video datasets, by discovering and distilling  
important cues about hands, objects, and interactions (albeit in a different viewpoint) that are relevant to egocentric activity during pre-training. 
To do this, we automatically assign third-person video instances with various egocentric pseudo-labels that span simple similarity scores 
to complex spatiotemporal attention maps, %
and then we introduce auxiliary losses that force our features to be predictive of these pseudo-labels. 
On the one hand, our approach retains the benefits of large-scale third-person video and the original action classification task to guide general video feature learning. On the other hand, we steer feature learning towards better egocentric features using automatically generated egocentric labels, as opposed to collecting manually labeled instances.

In the following sections, we first describe the traditional video pretraining framework (Sec~\ref{sec:background}) 
and how we incorporate our auxiliary loss terms into it (Sec~\ref{sec:pre-training}). Next we describe the three egocentric tasks we use, namely Ego-Score, Object-Score, and Interaction-Map (Sec~\ref{sec:tasks}). Finally, we present our full training and evaluation pipeline in Sec~\ref{sec:pipeline}. 






\subsection{Video model pre-training} \label{sec:background}




Video models benefit greatly from strong initializations. The standard procedure for training egocentric video models is thus to first pre-train models using large-scale third-person video datasets, and then fine-tune for a specific downstream task.

More formally, we are provided with a large-scale third-person (exocentric) video dataset $\mathcal{V}_{exo}$. In pre-training, each video instance $v \in \mathcal{V}_{exo}$ consists of $T$ frames $\{f_1, ..., f_T\}$ and an associated action label $y^{act}$. These frames are encoded into a series of $N$ spatiotemporal clip features $\{x_1, ..., x_N\}$, where $x_i \in \mathbb{R}^{c \times t \times h \times w}$, using a video encoder \emph{backbone} (e.g., a 3D CNN model). These features are then passed to a classifier \emph{head}, which spatiotemporally pools the feature and uses a linear classifier to generate the predicted action class $\hat{y}^{act}$. 
Predictions are typically generated for each clip and then averaged to generate video-level predictions. The network is trained to minimize the cross entropy loss $\mathbb{L}_{act}(y^{act}, \hat{y}^{act})$. See Fig~\ref{fig:framework} (left panel). 

Once pre-trained, the backbone weights are retained, the head is replaced with a task-specific classifier, and the new network is trained with instances from a target egocentric dataset $\mathcal{V}_{ego}$ to predict egocentric video labels.

\subsection{Ego-Exo pre-training} \label{sec:pre-training}
Third-person pre-training alone results in strong, general-purpose video features.  However, it ignores important egocentric signals 
and introduces a domain gap that limits its utility for downstream egocentric tasks. 
We introduce auxiliary egocentric task losses to overcome 
this gap.

Specifically, along with datasets $\mathcal{V}_{exo}$ and $\mathcal{V}_{ego}$, we assume access to off-the-shelf 
video models that address a set of egocentric video understanding tasks. 
For each task $\tau$, the model $M^\tau$ takes as input a video (as either frames or clips) and generates predicted labels $y^\tau$. We use these pre-trained models to associate egocentric \emph{pseudo-labels} to the third-person video instances in $\mathcal{V}_{exo}$.  We stress that the videos in $\mathcal{V}_{exo}$ are \emph{not} manually labeled for any task $\tau$.

We introduce 
a task-specific head $H^\tau$ for each task that is trained to approximate these pseudo-labels for each video instance, leading to an auxiliary loss term $\mathbb{L}_{\tau}(H^\tau(v), y^\tau)$. Each head is trained to approximate the response of an egocentric video model when applied to a third-person video instance, and thus can be seen as a knowledge-distillation mechanism 
that distills 
information from the egocentric tasks into the video encoder model.
The final pre-training objective is the combination of the action classification loss $\mathbb{L}_{act}$ and each of the auxiliary loss terms $\mathbb{L}_{\tau}$. See Fig~\ref{fig:framework} for our full framework.

Note that these pseudo-labels vary in structure and semantics, ranging from scalar scores (e.g., to characterize how \emph{egocentric-like} a third-person video is), categorical labels (e.g., to identify the manipulated objects in video) and spatiotemporal attention maps (e.g., to characterize hand-object interaction regions). Moreover, these labels are egocentric-specific, but they are \emph{automatically} generated for third-person video instances.  This diverse combination 
leads to robust feature learning for egocentric video, as our experiments will show. 
Once pre-trained, we can retain our \emph{enhanced} backbone weights to fine-tune on an egocentric video task using data from $\mathcal{V}_{ego}$.


\subsection{Auxiliary egocentric tasks}\label{sec:tasks}
Next, we describe each task we use, how we source $M^\tau$ and pseudo-labels $y^\tau$, the 
loss terms $\mathbb{L}_{\tau}$, and their relevance to egocentric feature learning. Note that no egocentric activity labels are used for the task models, and each task model is applied to third-person video instances in $\mathcal{V}_{exo}$.

\vspace{-3mm}
\paragraph{Ego-Score: Discriminating ego videos.}
A good egocentric video representation should be able to capture the underlying differences between first- and third-person videos, to discriminate between the two viewpoints. Based on this motivation, we design
an \textit{Ego-Score} task $\tau^{ego}$ 
to characterize the egocentricity likelihood of the video. 


For this, we train a binary \emph{ego-classifier} $M^{ego}$ on the Charades-Ego dataset~\cite{charades-ego}, which has both egocentric and third-person videos of indoor 
activities involving object interactions. 
 While the dataset offers paired instances showing the same activity from two views, our method does not use this pairing information or egocentric activity labels. 
It uses only the binary labels indicating if a sample is egocentric or exocentric. \YL{Please see Supp.~for more training details and an ablation study about the 
pairing information.}

We use this trained classifier to estimate the real-valued pseudo task-labels $y^{ego}$ for each video in our pre-training framework described in Sec~\ref{sec:pre-training}. We sample multiple clips from the same video and average their score to generate a video-level label. 
Formally, for a video $v$ with $N$ clips $\{x_1, ..., x_N \}$ we generate scores:
\vspace{-0.5mm}
\begin{equation}
y_i^{ego}(v) = \frac{\exp(\frac{1}{N\beta} \sum_{n} z^{ego}_i(x_n))}{\sum_{j} \exp(\frac{1}{N\beta} \sum_{n}{z^{ego}_j(x_n))}}, \label{eqn:yego}
\end{equation}
where $\beta$ is a scalar temperature parameter, $z^{ego}_i(x_n)$ is the predicted logits from the ego-classifier $M^{ego}$, and $i \in \{0, 1\}$ is the class label. 

 
Third-person videos display various egocentric cues, resulting in a broad distribution of values for Ego-Score, despite sharing the same viewpoint (details in Supp).
This score is used as the soft target in the auxiliary task loss, which we predict using a video classification head $H^{ego}$: 
\begin{equation}
    \mathbb{L}_{ego}(x) = -\sum_{i} y^{ego}_i(v)\ \log (H^{ego}_i(x)).
\vspace{-4mm}
\end{equation}


\vspace{-2mm}
\paragraph{Object-Score: Finding interactive objects.}
In egocentric videos, 
interactions with objects are often central, as evident 
in popular egocentric video datasets~\cite{charades-ego,epic,li2018eye}. 
Motivated by this, we designate an \textit{Object-Score} task $\tau^{obj}$ for each video that encourages video representations to be predictive of manipulated objects.

Rather than require 
ground-truth object labels for third-person videos, we propose a simple solution that directly uses an off-the-shelf object recognition model $M^{obj}$ trained on ImageNet~\cite{deng2009imagenet} 
to describe objects in the video. Formally, for a video $v$ with frames $\{f_1, ..., f_T\}$ 
we average the predicted logits from $M^{obj}$ 
across frames to generate the video-level Object-Score $y^{obj}_i(v)$:  
\begin{equation}
    y^{obj}_i(v) = \frac{\exp(\frac{1}{T\beta} \sum_{t} z^{obj}_i(f_t))}{\sum_{j} \exp(\frac{1}{T\beta} \sum_{t}{z^{obj}_j(f_t))}}, \label{eqn:yobj}
\end{equation}
where $z^{obj}_i(f_t)$ is predicted logits for the $i^{th}$ class from the recognition model, and $\beta$ is the temperature parameter. 

Similar to the Ego-Score, we introduce 
a knowledge-distillation loss during pre-training to make the video model predictive of the Object-Score using a module $H^{obj}$:
\begin{equation}
    \mathbb{L}_{obj}(x) = -\sum_{i} y^{obj}_i(v) \log H^{obj}_i(x).
\vspace{-3mm}
\end{equation}




\paragraph{Interaction-Map: Discovering hand interaction regions.}
The 
Object-Score attempts to describe the interactive objects. Here we explicitly 
focus on the spatiotemporal regions of interactions in videos. 
Prior work shows it is possible to recognize a camera wearer's actions by attending to only a small region around the gaze point~\cite{li2018eye}, as gaze often focuses on hand-object manipulation.
Motivated by this, 
we introduce an \textit{Interaction-Map} task $\tau^{int}$ to learn features that are
predictive of these important spatiotemporal hand-object interaction regions in videos. 

\begin{figure}[t]
\centering
\includegraphics[width=\columnwidth]{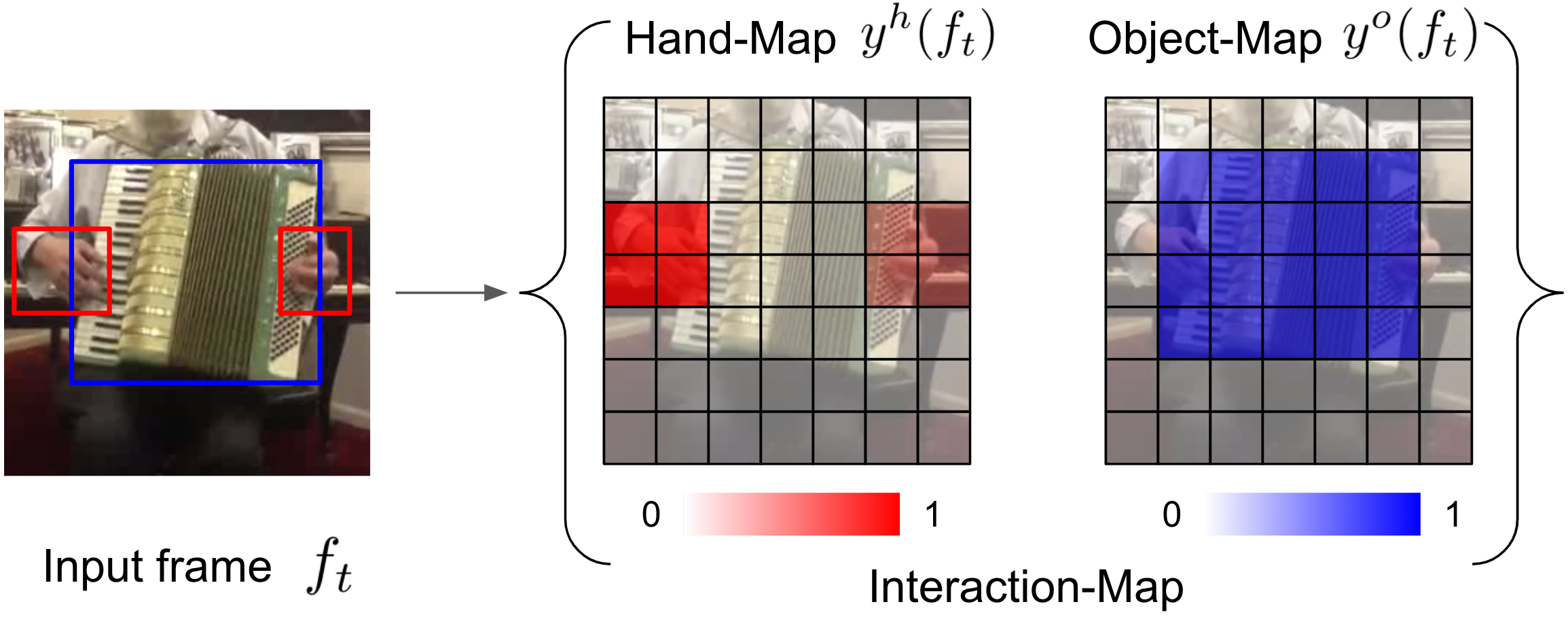}
\vspace{-7mm}
\caption{\textbf{Interaction-Map construction.} Soft-labels for hand-map and object-map are generated from detected bounding boxes according to Eqn~\ref{eq:bbox}.
}
\vspace{-5mm}
\label{fig:interaction-map}
\end{figure}

We adopt an off-the-shelf hand-object detector~\cite{shan2020understanding} $M^{int}$ to detect hands and interacting objects. 
For each frame $f_t$ in a video, the hand detector predicts a set of bounding-box coordinates and associated confidence scores $\mathcal{B}^t = \{(b^h, s^h)\}$ for detected hands. These bounding boxes are scaled to $h \times w$---the spatial dimensions of the video clip feature. 
We then generate a $t \times h \times w$ spatiotemporal hand-map $y^{h}$ where the score for each grid-cell $y^{h}_{i, j} (f_t)$ at each time-step is calculated by its overlap with the detected hands at time $t$: 
\begin{align}
\vspace{-1mm}
     y^{h}_{i, j} (f_t) = 
    \begin{cases}
    \max_{k \in \mathcal{B}^t_{i, j}}\{s^h_k\} & \text{if} ~~~~ \mathcal{B}^t_{i, j} \neq \emptyset, \\
    0 & \text{if} ~~~~ \mathcal{B}^t_{i, j} = \emptyset,
    \end{cases}  
    \label{eq:bbox}
\vspace{-1mm}
\end{align}
where $\mathcal{B}^t_{i, j}$ 
is the set of predicted bounding boxes that overlap with the $(i, j)$-th grid-cell at time $t$. The full hand-map label $y^h$ is formed by concatenating the per-frame labels over time.
We generate the corresponding object-map $y^o$ analogously. See Fig~\ref{fig:interaction-map} for an illustrative example.

We use the hand-map and object-map as the Interaction-Map pseudo-labels for the third-person videos during pre-training. We introduce two Interaction-Map prediction heads, $H^h$ and $H^o$, 
to directly predict the $t \times h \times w$ Interaction-Map labels from clip features using a 3D convolution head:
\begin{equation}
\vspace{-1mm}
\begin{split}
    \mathbb{L}_{int}(x) = - \sum_{i, j} \Bigg( \Bigg.
    \sum_{k} y^{h}(f_k) \log [H^{h}(x)]_k \\
    + y^{o}(f_k) \log [H^{o}(x)]_k 
    \Bigg. \Bigg)_{i, j},
\end{split} \label{eq:int-map}
\vspace{-3mm}
\end{equation}
where $f_k$ is $k$-th frame from training clip $x$, and $H^{h}(x)$ and $H^{o}(x)$ are the predicted hand-map and object-map scores. 

\cc{We predict Interaction-Maps instead of directly predicting bounding boxes 
via standard detection networks~\cite{ren2015faster} for two reasons. 
First, detection architectures are not directly compatible with standard video 
backbones---they 
typically utilize specialized backbones 
and work well only with 
high resolution inputs. 
Second, predicting scores on 
a feature map 
is more aligned with our ultimate goal to improve the feature representation for egocentric video tasks, rather than train a precise detection model.  }


\subsection{Ego-Exo training and evaluation}\label{sec:pipeline}
The three proposed ego-specific auxiliary tasks are 
combined together during the pre-training procedure to construct the final training loss:
\vspace{-1mm}
\begin{equation}
\vspace{-1mm}
\begin{split}
     \mathbb{L}(x) = \mathbb{L}_{act}(x) &  + w_{ego} * \mathbb{L}_{ego}(x) \\
            +  w_{obj} & * \mathbb{L}_{obj}(x) + w_{int} * \mathbb{L}_{int}(x),
\end{split} \label{eqn:loss}
\end{equation} 
where $\mathbb{L}_{act}$ is the standard cross-entropy loss for third-person action recognition, and $w_{ego}$, $w_{obj}$ and $w_{int}$ are the corresponding loss weights for the three auxiliary tasks, selected via cross-validation 
on downstream tasks. 

Note that third-person video instances without hand-object interactions or salient interactive objects still contribute to the auxiliary loss terms, and are not ignored. Our distillation models 
approximate the \emph{responses} of the pre-trained egocentric models as soft-targets instead of hard labels, offering valuable information about perceived egocentric cues, whether positive or negative for the actual label. 


Training with our auxiliary losses results in features that are more suitable for downstream egocentric tasks, but it does not modify the network architecture itself. Consequently, after pre-training, our model can be directly used as a drop-in replacement for traditional video encoders, and it can be applied to various egocentric video tasks. 



\section{Experiments}~\label{sec:exp}

\vspace{-10mm}
\paragraph{Datasets.} 
Our experiments use the following datasets.
\begin{itemize}[leftmargin=*]
\vspace{-1mm}
\setlength\itemsep{0em}
\item \textbf{Kinetics-400}~\cite{kinetics} is a popular third-person video dataset containing $\sim$300k videos 
and spanning 400 human action classes. 
We use this dataset to pre-train all our models.
\item \textbf{Charades-Ego}~\cite{charades-ego} has $\sim$68k 
instances spanning 157 activity classes. 
Each instance is a pair of videos corresponding to the same activity, recorded in the first and third-person perspective.  Our method does not require this pairing, and succeeds even if no pairs exist (Supp.). 

\item \textbf{EPIC-Kitchens}~\cite{epic} is an egocentric video dataset with videos of non-scripted daily activities in kitchens. It contains 55 hours of videos consisting of 39k action segments, annotated for interactions spanning 352 objects and 125 verbs. \textbf{EPIC-Kitchens-100}~\cite{epic-100} extends this to 100 hours and 90k action segments, and is currently the largest annotated egocentric video dataset. 
\end{itemize}

Due to its large scale and diverse coverage of actions, Kinetics has widely been adopted as the standard dataset for pre-training both first- and third-person video models~\cite{gu2018ava,slowfast,epic-fusion,epic-100}. 
EPIC-Kitchens and Charades-Ego are two large and challenging egocentric video datasets that are the subject of recent benchmarks and challenges.

\paragraph{Evaluation metrics.}
We pre-train all models on Kinetics, and fine-tune on Charades-Ego (first-person only) and EPIC-Kitchens for activity recognition. Following standard practice, 
we report mean average precision (mAP) for Charades-Ego 
and top-1 and top-5 accuracy for EPIC. 

\paragraph{Implementation details.} 
We build our Ego-Exo framework on top of PySlowFast~\cite{pyslowfast} and use SlowFast~\cite{slowfast} video models 
as backbones with $8$ input frames and stride $8$. 
We use a Slow-only ResNet50 architecture for all ablation experiments, and  a SlowFast ResNet50/101 architecture for final results. 

For our distillation heads $H^{ego}$ and $H^{obj}$ (Sec~\ref{sec:tasks}), we use a spatiotemporal pooling layer, followed by a linear classifier. We implement our Interaction-Map heads $H^{h}$ and $H^{o}$ as two 3D conv layers with kernel sizes $1{\mkern-2mu\times\mkern-2mu}3{\mkern-2mu\times\mkern-2mu}3$ and $1{\mkern-2mu\times\mkern-2mu}1{\mkern-2mu\times\mkern-2mu}1$, and ReLU activation. 

For our combined loss function (Eqn~\ref{eqn:loss})
, we set the 
loss weights $w_{ego}$, $w_{obj}$ and $w_{int}$ to $0.1, 0.5, 1.0$ respectively through cross-validation on the EPIC-Kitchens validation set (cross-validation on Charades-Ego suggested
similar weights).
The temperature parameter $\beta$ in Eqn~\ref{eqn:yego} and Eqn~\ref{eqn:yobj} is set to 1. 
Training schedule and optimization details can be found in Supp.

\subsection{Ego-Exo pre-training}

We compare our pre-training strategy to these methods: 
\begin{itemize}[leftmargin=*]
\vspace{-1mm}
\setlength\itemsep{-0.3em} 
    \item \textbf{Scratch} does not benefit from any pre-training. It is randomly initialized and directly fine-tuned on the target egocentric dataset.
    \item \textbf{Third-only} is pre-trained for activity labels on Kinetics 400~\cite{kinetics}. This represents the status-quo pre-training strategy for current video models.
    \item \textbf{First-only} is pre-trained for verb/noun labels on EPIC-Kitchens-100~\cite{epic-100}, the largest publicly available egocentric dataset. 
    \item \textbf{Domain-adapt} 
    introduces a domain adaptation loss derived from gradients of a classifier trained to distinguish between first- and third-person video instances~\cite{ganin2016domain}.
    This strategy has been used in recent work to learn domain invariant features for third-person vs.~drone footage~\cite{choi2020unsupervised}. 
    \item \textbf{Joint-embed} uses paired first- and third-person video data from Charades-Ego to learn viewpoint-invariant video models via standard triplet embedding losses~\cite{sigurdsson2018actor}. We first pre-train this model with Kinetics to ensure that the model benefits from large-scale pre-training. 
    \item \textbf{Ego-Exo} is pre-trained on Kinetics-400, but additionally incorporates  
    the three auxiliary egocentric tasks (Sec~\ref{sec:tasks}) together with the original action classification loss, to learn egocentric-specific features during pre-training. 
\vspace{-1mm}
\end{itemize}
For this experiment, all models share the same backbone architecture (Slow-only, ResNet-50)
and only the pre-training strategy is varied to ensure fair comparisons. \emph{Domain Adapt} uses additional unlabeled egocentric data during pre-training, but from the same target dataset that the model will have access to during fine-tuning. \emph{Joint-embed} uses paired egocentric and third-person data, an advantage that the other methods do not have, but offers insight into performance in this setting.  Only \emph{First-only} has access to ego-videos labeled for actions during pre-training.  


\begin{table}[t]
\small
	\begin{center}
	\setlength{\tabcolsep}{3.5pt}
	\begin{tabular}{l|c|cc|cc}
		    &  C-Ego & \multicolumn{2}{c|}{EPIC verbs} & \multicolumn{2}{c}{EPIC nouns} \\
		\cline{2-6}
		Methods	&  mAP  & top-1 & top-5 & top-1 & top-5	\\
		\hline
        Scratch  & 8.50 & 55.62 & 86.10 & 38.35 & 62.39 \\
        First-only & 11.73  & -- & -- & -- & --	  \\
        Third-only & 24.69 & 61.19 & 87.49 & 46.18 & 69.72 \\
        Domain-adapt~\cite{ganin2016domain} & 23.62 & 61.27 & 87.49 & 45.93 & 68.73 	  \\
        Joint-embed~\cite{sigurdsson2018actor} & - & 61.26 & 87.17 & 46.55 & 68.97	  \\
        Ego-Exo       & \textbf{26.23} & \textbf{62.83} & \textbf{87.63} & \textbf{48.15} & \textbf{70.28} \\
		\hline				
	\end{tabular}
	\end{center}
\vspace{-5mm}
\caption{\textbf{Ego-Exo vs.~alternate pre-training methods.} 
Our Ego-Exo pre-training results in best performance.
Note that we do not evaluate \emph{First-only} and \emph{Joint-embed} on the datasets they were pre-trained on (EPIC and Charades-Ego respectively). Values are averaged over 3 runs. 
}
\vspace{-1mm}
\label{tb:different_baseline}
\end{table}

\begin{table}[t]
\centering
\resizebox{\columnwidth}{!}{
	\setlength{\tabcolsep}{3pt}
	\begin{tabular}{l|ccc|c|cc|cc}
		    & & & & C-Ego & \multicolumn{2}{c|}{EPIC verbs} & \multicolumn{2}{c}{EPIC nouns} \\
		\cline{5-9}
 		Methods	& $\tau^{ego}$ & $\tau^{obj}$ & $\tau^{h+o}$ & mAP  & top-1 & top-5 & top-1 & top-5	\\
		\hline
        Third-only & \xmark & \xmark & \xmark & 24.69 & 61.19 & 87.49 & 46.18 & 69.72\\
        \hline
        \multirow{5}{*}{Ego-Exo} & \cmark & \xmark & \xmark & 25.01 & 62.22 & 87.78 & 46.26 & 68.76 \\
            &  \xmark & \cmark & \xmark & 25.49 & 61.65 & 87.57 & 46.27 & 69.52\\
            &  \xmark & \xmark & \cmark & 25.91 & 62.55 & \textbf{88.50} & 47.71 & 69.62\\
            &  \cmark & \cmark & \cmark & \textbf{26.23} & \textbf{62.83} & 87.63 & \textbf{48.15} & \textbf{70.28} 	\\
		\hline				
	\end{tabular}
}
\vspace{-2mm}
\caption{\textbf{Auxiliary task ablation study.} Distilling knowledge from all three egocentric tasks results in the best performing pre-trained model. Values are averaged over 3 runs.
}
\vspace{-3mm}
\label{tb:different_aux}
\end{table}

Table~\ref{tb:different_baseline} shows the validation performance of different pre-training strategies. 
\emph{Third-only} benefits from strong initialization from large-scale video pre-training and greatly outperforms models trained from \emph{Scratch}.
\emph{First-only} performs very poorly despite being 
pre-trained on the largest available egocentric dataset, 
indicating that increasing scale alone is not sufficient---the diversity of scenes and activities in 
third-person data plays a significant role in feature learning as well. 
\emph{Domain-adapt} and \emph{Joint-embed} both learn viewpoint invariant features using additional unlabeled egocentric data. 
However, the large domain gap and small scale of the paired dataset limit improvements over \emph{Third-only}. 
Our Ego-Exo method achieves the best 
results on both Charades-Ego and EPIC-Kitchens. The consistent improvements (+1.54\% mAP on Charades-Ego, and +1.64\%/+1.97\% on EPIC verbs/nouns) over Third-only demonstrate the effectiveness of our proposed auxiliary egocentric tasks during pre-training.  This is a key result showing the impact of our idea. 

\begin{figure*}[t!]
\centering
\includegraphics[width=2.0\columnwidth]{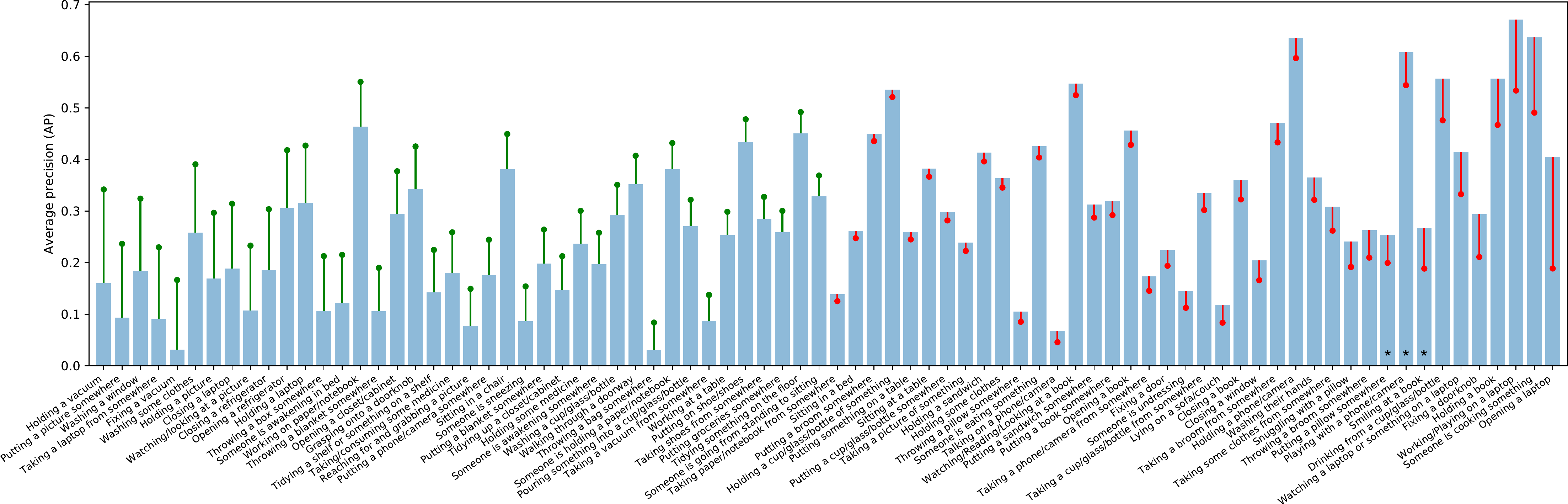}
\vspace{-2mm}
\caption{\textbf{Class-level performance on Charades-Ego.} Our method significantly improves on several classes that focus on active object manipulations (green lines), and performs only marginally worse across most under-performing classes (red lines). 35 most/least improved classes (out of all 157 classes) are shown.
}
\vspace{-3mm}
\label{fig:classwise}
\end{figure*}

Fig~\ref{fig:classwise} shows a class-wise breakdown of performance on Charades-Ego compared to the \emph{Third-only} baseline. Our pre-training strategy results in larger improvements on classes that involve active object manipulations. 

\subsection{Ablation studies}

\begin{table}[t]
\small
\centering
\resizebox{\columnwidth}{!}{
	\setlength{\tabcolsep}{3.5pt}
	\begin{tabular}{l|c|cc|cc}
		    &  C-Ego & \multicolumn{2}{c|}{EPIC verbs} & \multicolumn{2}{c}{EPIC nouns} \\
		\cline{2-6}
		Methods	&  mAP  & top-1 & top-5 & top-1 & top-5	\\
		\hline
        Third-only  &  24.69 & 61.19 & 87.49 & 46.18 & 69.72  \\
        Third-only +aux &  25.00 & 62.36 & 87.72 & 46.59 & 68.33 \\
        {$\Delta$} & \textit{+0.31} & \textit{+1.17} & \textit{+0.23} & \textbf{\textit{+0.42}} & \textit{-1.39} \\
        \hline
        Ego-Exo  & 26.23 & 62.83 & 87.63 & 48.15 & 70.28 \\
        Ego-Exo +aux & 27.47 & 64.26 & 88.45 & 48.39 & 70.68\\
        {$\Delta$} &  \textbf{\textit{+1.24}} &  \textbf{\textit{+1.43}} &  \textbf{\textit{+0.82}} & \textit{+0.24} & \textbf{\textit{+0.40}} \\
		\hline				
	\end{tabular}
}
\vspace{-2mm}
\caption{\textbf{Effect of Ego-Exo losses during fine-tuning.} Adding distillation losses during fine-tuning improves performance for both models, and results in a larger performance gain for our Ego-Exo pre-trained models. \cc{Values are averaged over 3 runs.}
}
\vspace{-3mm}
\label{tb:add_loss_to_ego}
\end{table}

\paragraph{Impact of auxiliary ego-tasks.}
We next analyze the impact of each auxiliary egocentric task in our Ego-Exo framework. 
As shown in Table~\ref{tb:different_aux}, adding the \emph{Ego-Score} task improves performance 
on both EPIC-Kitchens tasks, while adding \emph{Object-Score} and \emph{Interaction-Map} consistently improves all results. 
This reveals that despite varying structure and semantics, these scores capture important underlying egocentric information to complement 
third-person pre-training, and further boost performance when used together. 

\begin{figure}[t]
\centering
\includegraphics[width=\columnwidth]{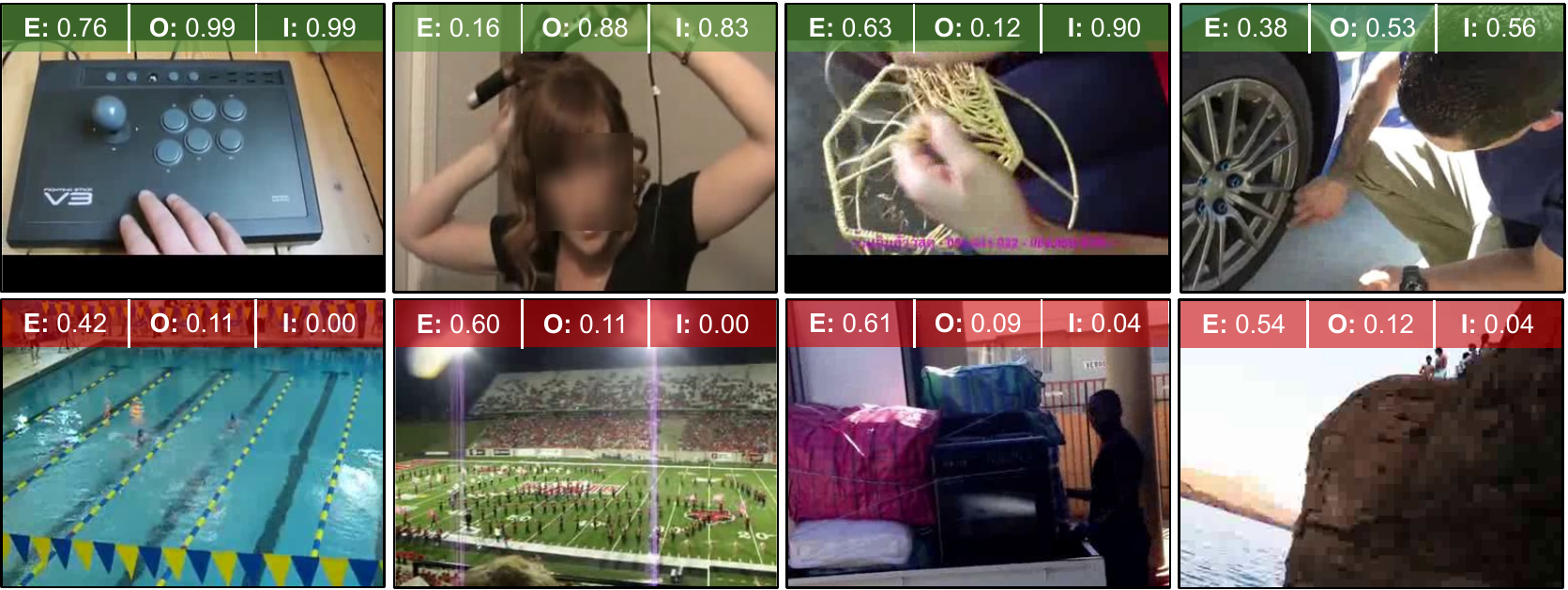}
\vspace{-4mm}
\caption{\textbf{Kinetics instances sorted by Ego-Exo scores.} Our task scores are maximum for videos that prominently feature hands/objects in view (top row), and minimum for scenes devoid of human-centered activity (bottom row). 
}
 \vspace{-3mm}
\label{fig:retrieval-swatch}
\end{figure}

Fig~\ref{fig:retrieval-swatch} shows instances from Kinetics based on our auxiliary pseudo-label scores combined with the weights in Eqn~\ref{eqn:loss}. 
Our score is highest for object-interaction heavy activities (e.g., top row: knitting, changing a tire), while it is low for videos of broader scene-level activities (e.g., bottom row: sporting events). Note that these videos are not in the egocentric viewpoint---they are largely third-person videos from static cameras, but are \emph{ego-like} in that they prominently highlight important features of egocentric activity (e.g. hands, object interactions).


\vspace{-2mm}
\paragraph{Adding auxiliary ego-tasks during fine-tuning.} 
Our auxiliary losses may also be added \emph{after} pre-training, for fine-tuning downstream egocentric models similar to prior semi-supervised learning work~\cite{chen2020big}. 
We re-introduce our Interaction-Map loss $\mathbb{L}_{int}$ (Eqn~\ref{eq:int-map}) 
for downstream egocentric training. We do not include Ego-Score (which is trivially high for all videos) and Object-Score (which is subsumed in the interaction label for this setting) 
as their impact 
after pre-training was minimal. 


Table~\ref{tb:add_loss_to_ego} shows that while both the baseline and our method further improve by adding the auxiliary task during fine-tuning, 
our improvements (Ego-Exo + aux) are larger, especially on Charades-Ego. 
This is likely because 
our distillation heads benefit from training to detect hands and objects in large-scale third-person video prior to fine-tuning for the same task on downstream ego-datasets.

\begin{table}[tb]
\small
\renewcommand\arraystretch{1.1}
	\begin{center}
	\begin{tabular}{l|c}
		Method	& mAP	\\
		\hline
		ActorObserverNet~\cite{sigurdsson2018actor} & 20.0 \\
		SSDA~\cite{choi2020unsupervised} & 23.1 \\
        I3D~\cite{choi2020unsupervised} & 25.8 \\
	    \hline
        SlowFast~\cite{slowfast} & 25.93 \\
        Ego-Exo & 28.04 \\
        Ego-Exo* & 29.19 \\
        Ego-Exo*-R101 & \textbf{30.13} \\
        \hline
	\end{tabular}
	\end{center}
\vspace{-5mm}
\caption{\textbf{Comparison to prior work on Charades-Ego.} Despite having no access to paired egocentric data, our model outperforms specialized joint-embedding and domain adaptation based methods.  
}
\vspace{-3mm}
\label{tb:charades-ego-results}
\end{table}

\subsection{Comparison with state-of-the-art}

\begin{table*}[tb]
\small
	\begin{center}
	\setlength{\tabcolsep}{4.5pt}
	\begin{tabular}{c|l|cccccc|ccc|ccc}
	   & & \multicolumn{6}{c|}{Overall} & \multicolumn{3}{c|}{Unseen Participants} & \multicolumn{3}{c}{Tail Classes} \\
	   \cline{3-14}
	  &  & \multicolumn{3}{c|}{top-1} & \multicolumn{3}{c|}{top-5} &	\multicolumn{3}{c|}{top-1} & \multicolumn{3}{c}{top-1} \\
	   \cline{3-14}
	&	Methods &  verb & noun & \multicolumn{1}{c|}{actions} & verb & noun & actions & verb & noun & actions & verb & noun & actions \\
	
		\hline
	\color{darkgray}{{w/ audio}} &	\color{darkgray}{{Epic-Fusion~\cite{epic-fusion}}} & \color{darkgray}{{62.40}} & \color{darkgray}{{46.50}} & \color{darkgray}{{35.11}} & \color{darkgray}{{88.74}} & \color{darkgray}{{72.24}} & \color{darkgray}{{55.17}} & \color{darkgray}{{56.57}} & \color{darkgray}{{41.78}} & \color{darkgray}{{29.46}} & \color{darkgray}{{29.41}} & \color{darkgray}{{18.55}} & \color{darkgray}{{13.47}} \\
		\Xhline{2\arrayrulewidth}
	\multirow{7}{*}{w/o audio} &	TSN fusion~\cite{epic} &  58.43 & 46.54 & 32.79 & 87.27 & 72.49 & 53.12 & 52.04 & 42.09 & 26.30 & 24.76 & 14.91 & 10.41 \\
    	& TRN~\cite{zhou2018temporal} & 62.56 & 45.70 & 34.41 & 88.24 & 71.37 & 54.65 & 57.49 & 40.85 & 28.69 & 27.24 & 13.42 & 11.20 \\
		& TSM~\cite{lin2019tsm} & 65.51 & 48.48 & 37.58 & \textbf{89.39} & 73.46 & 58.04 & 59.66 & 43.16 & 30.41 & 29.76 & 15.84 & 13.15 \\
        \cline{2-14}
        & SlowFast~\cite{slowfast} & 63.89 & 49.66 & 37.42 & 88.71 & 74.99 & 58.17 & 57.37 & 44.31 & 29.71 & 33.57 & 22.57 & 16.55 \\
        & Ego-Exo & 64.08 & 50.59 & 38.38 & 88.57 & 74.82 & 59.04 & 57.42 & 46.18 & 31.41 & 32.87 & \textbf{22.98} & 16.30 \\
        & Ego-Exo* & 65.02 & \textbf{51.74} & 39.52 & 89.26 & 75.95 & 60.07 & 58.86 & \textbf{47.01} & 32.36 & 33.68 & 22.35 & 16.30 \\
        & Ego-Exo*-R101 & \textbf{66.07} & 51.51 & \textbf{39.98} & \textbf{89.39} & \textbf{76.31} & \textbf{60.68} & \textbf{59.83} & 45.50 & \textbf{32.63} & \textbf{33.92} & 22.91 & \textbf{16.96} \\
        \hline
	\end{tabular}
	\end{center}
\vspace{-5mm}
\caption{\textbf{Comparison on EPIC-Kitchens-100 action recognition test set.} Our method 
is best in all categories.
}
\vspace{-4.5mm}
\label{tb:epic-100-results}
\end{table*}

Finally, we compare our method with state-of-the-art models, 
many of which use additional modalities (flow, audio) compared to our RGB-only models. 
We include three competitive variants of our model using SlowFast~\cite{slowfast} backbones: (1) \textbf{Ego-Exo} uses a ResNet50 backbone; (2) \textbf{Ego-Exo*} additionally incorporates our auxiliary distillation loss during fine-tuning.\footnote{Same as Ego-Exo + aux in Table~\ref{tb:add_loss_to_ego}, but here with a SlowFast backbone} (3) \textbf{Ego-Exo*-R101} 
further uses a ResNet-101 backbone.

\vspace{-1mm}
\paragraph{Charades-Ego.}
Table~\ref{tb:charades-ego-results} compares our Ego-Exo method with 
existing methods. Our Ego-Exo and Ego-Exo* yield state of the art accuracy, improving performance over the strongest baseline by +2.11\% and +3.26\% mAP. We  observe large performance gains over prior work, 
including ActorObserverNet~\cite{sigurdsson2018actor} and SSDA~\cite{choi2020unsupervised}, which use joint-embedding or domain adaptation approaches to transfer third-person video features to the first-person domain. In addition, unlike the competing methods, our method does not require any egocentric data which is paired or shared category labels with third-person data during pre-training.

\vspace{-3mm}
\paragraph{EPIC-Kitchens.} 
Table~\ref{tb:epic-55-results} compares our method to state-of-the-art models on the EPIC-Kitchens test set. 
Ego-Exo and Ego-Exo* 
consistently improve over SlowFast (which shares the same backbone architecture) 
for all 
categories on both seen and unseen test sets.
Epic-Fusion~\cite{epic-fusion} uses additional optical flow and audio modalities together with RGB, yet 
Ego-Exo outperforms it on the top-1 metric for all categories. 
AVSlowFast~\cite{xiao2020audiovisual} also utilizes audio, but is outperformed by our model with the same backbone (Ego-Exo*-R101) 
on the S1 test set.\footnote{Table~\ref{tb:epic-55-results} compares existing methods under a controlled setting: using a single model with RGB or RGB+audio as input, and only Kinetics/ImageNet for pre-training.  Other reported results on the competition page may use extra modalities, larger pre-training datasets, or model ensemble schemes (e.g. the top ranking method ensembles 8 models).}
On 
\textbf{EPIC-Kitchen-100}~\cite{epic-100}, as shown in Table~\ref{tb:epic-100-results}, Ego-Exo consistently improves over the SlowFast baseline on all evaluation metrics, and Ego-Exo*-R101 outperforms \emph{all} existing 
state-of-the-art methods.

\begin{table}[t]
\centering
\resizebox{\columnwidth}{!}{
\setlength{\tabcolsep}{3pt}
	\begin{tabular}{c|l|cc|cc|cc}
		&	& \multicolumn{2}{c|}{verbs} & \multicolumn{2}{c|}{nouns} & \multicolumn{2}{c}{actions}\\
		\cline{2-8}
	\textbf{S1 (seen)}	& Methods	&   top-1 & top-5 & top-1 & top-5 & top-1 & top-5	\\
		\hline
	 \color{darkgray}{{\multirow{2}{*}{w/ audio}}}	& \color{darkgray}{{Epic-Fusion}~\cite{epic-fusion}} & \color{darkgray}{{64.75}} & \color{darkgray}{{\textbf{90.70}}} & \color{darkgray}{{46.03}} & \color{darkgray}{{71.34}}  & \color{darkgray}{{34.80}} & \color{darkgray}{{56.65}} \\
	&	\color{darkgray}{{AVSF-R101}~\cite{xiao2020audiovisual}} & \color{darkgray}{\textbf{{65.70}}} & \color{darkgray}{{89.50}}  & \color{darkgray}{\textbf{{46.40}}}  & \color{darkgray}{{\textbf{71.70}}}  & \color{darkgray}{\textbf{{35.90}}}  & \color{darkgray}{{\textbf{57.80}}} \\
		\Xhline{2\arrayrulewidth}
	\multirow{6}{*}{w/o audio} &	TSN fusion~\cite{epic} & 48.23 & 84.09 & 36.71 & 62.32 & 20.54 & 39.79  \\
	&	RU-LSTM~\cite{furnari2019would} & 56.93 & 85.68 & 43.05 & 67.12 & 33.06 & 55.32 \\
		\cline{2-8}
    &    SlowFast~\cite{slowfast} & 64.57 &	89.67&	45.89&	69.50&	34.67&	54.47 \\
    &    Ego-Exo &  65.84&	89.91&	46.68&	70.30&	36.00&	54.90\\
     &    Ego-Exo* & \textbf{66.19} & 90.11 & 47.21 & 70.08 & 37.04 & 55.91  \\
    &    Ego-Exo*-R101 & 65.97 & \textbf{90.32} & \textbf{47.99} & \textbf{70.72} & \textbf{37.09} & \textbf{56.32} \\
        \hline
        \hline
    \textbf{S2 (unseen)} & & & & & & &  \\
    \hline
	 \color{darkgray}{{\multirow{2}{*}{w/ audio}}} &    \color{darkgray}{{Epic-Fusion}~\cite{epic-fusion}} & \color{darkgray}{{52.69}} & \color{darkgray}{{79.93}} & \color{darkgray}{{27.86}} & \color{darkgray}{{53.87}} & \color{darkgray}{{19.06}} & \color{darkgray}{{36.54}} \\
	&	\color{darkgray}{{AVSF-R101}~\cite{xiao2020audiovisual}} & \color{darkgray}{{\textbf{55.80}}} & \color{darkgray}{{\textbf{81.70}}} & \color{darkgray}{{\textbf{32.70}}} & \color{darkgray}{{\textbf{58.90}}} & \color{darkgray}{{\textbf{24.00}}} & \color{darkgray}{{\textbf{43.20}}} \\
		\Xhline{2\arrayrulewidth}
	\multirow{6}{*}{w/o audio} &	TSN fusion~\cite{epic} &  39.40 & 74.29 & 22.70 & 45.72 & 10.89 & 25.26 \\
	&	RU-LSTM~\cite{furnari2019would} & 43.67 & 73.30 & 26.77 & 48.28 & 19.49 & 37.15 \\
        \cline{2-8}
    &    SlowFast~\cite{slowfast} & 53.91 &	80.81 &	30.15 &	55.48 &	21.58 &	37.56 \\
    &    Ego-Exo &  54.11 & 	80.37 &	30.83 & 57.08 &	21.20 & 38.34\\
    &    Ego-Exo* & 53.88 & 80.51 & 31.44 & 57.60 & 22.02 & 39.13 \\
    &    Ego-Exo*-R101 & \textbf{55.34} & \textbf{81.46} & \textbf{31.72} & \textbf{58.25} & \textbf{22.81} & \textbf{40.18} \\
        \hline
	\end{tabular}
}
\vspace{-2mm}
\caption{\textbf{Comparison to prior work on EPIC-Kitchens (test set).} 
Methods in \textcolor{darkgray}{gray} use additional audio modality information. Our method outperforms all methods that use consistent modalities in both settings, and is competitive with models that benefit from audio stream inputs. 
}
\vspace{-4mm}
\label{tb:epic-55-results}
\end{table}

\section{Conclusion}
We proposed a novel method to embed key egocentric signals into the traditional third-person video pre-training pipeline, so that models could benefit from both the scale and diversity of third-person video datasets, and create strong video representations for downstream egocentric understanding tasks.
Our experiments show the viability of our approach as a drop-in replacement for the standard Kinetics-pretrained video model, achieving state-of-the-art results on egocentric action recognition on Charades-Ego and EPIC-Kitchens-100. 
Future work could explore alternate distillation tasks and instance-specific distillation losses to maximize the impact of third-person data for training egocentric video models. 

\vspace{0.1in}
\noindent\textbf{Acknowledgments}: UT Austin is supported in part by the NSF AI Institute and FB Cognitive Science Consortium.

{\small
\bibliographystyle{ieee_fullname}
\bibliography{egbib}
}

\newpage
\clearpage

\setcounter{section}{0}
\setcounter{figure}{0}
\setcounter{table}{0}
\renewcommand{\thesection}{S\arabic{section}}
\renewcommand{\thetable}{S\arabic{table}}
\renewcommand{\thefigure}{S\arabic{figure}}




\section*{Supplementary Material}

This section contains supplementary material to support the main paper text. The contents include:

\begin{itemize}[leftmargin=*]
\itemsep0em 
    \item (\S\ref{sec:supp_egomodel}) Implementation details for three pre-trained egocentric models $M^\tau$ from Sec.~\ref{sec:tasks}.
    \item (\S\ref{sec:supp_pretrain}) Implementation details for Kinetics pre-training presented in Sec.~\ref{sec:background}.
    \item (\S\ref{sec:supp_finetune}) Implementation details for fine-tuning on downstream egocentric datasets.   
    \item (\S\ref{sec:supp_results}) Additional results 
    on EPIC-Kitchens, Charades-Ego and Something-Something v2 datasets.
    \item (\S\ref{sec:supp_ablation}) Additional ablation studies, including ablations of the Interaction-map model, using $M^\tau$ as pre-trained models, appending features from $M^\tau$, the impact of egocentric dataset scale on model performance, implicit pairing information in Ego-Scores and effect of using different  backbones for Ego-Exo.
    \item (\S\ref{sec:supp_qualitative}) Additional qualitative results, including distribution of Ego-Score over Kinetics, additional qualitative examples and class-wise breakdown of improvements for auxiliary tasks $M^{\tau}$. 
    \item \textbf{Supplementary video.} A demonstration video shows animated version of video clips for the qualitative examples in \S\ref{sec:supp_qualitative}.
\end{itemize}

\section{Details: Pre-trained egocentric models $M^{\tau}$} \label{sec:supp_egomodel}
We provide additional implementation details for the task models used in Auxiliary egocentric tasks from Sec.~\ref{sec:tasks}.

\paragraph{Ego-Classifier $M^{ego}$ for Ego-Score.} 
We use a Slow-only model~\cite{slowfast} with a ResNet-50 backbone as the ego-classifier $M^{ego}$. Then, we train $M^{ego}$ on the Charades-Ego dataset~\cite{charades-ego} in which each instance is assigned with a binary label indicating if it is egocentric or exocentric. We take a Kinetics-pretrained model as initialization and train with 8 GPUs in 100 epochs. We adopt a cosine schedule for learning rate decaying with a base learning rate as 0.01 and the mini-batch size is 8 clips per GPU. To generate pseudo-labels for Kinetic videos, we sample $N=2$ clips for each video and generate our Ego-Score using Eqn~\ref{eqn:yego}.

\paragraph{Object recognition model $M^{obj}$ for Object-Score.} We directly use an off-the-shelf object recognition model trained on ImageNet as $M^{obj}$. Specifically, we use a standard ResNet-152 network from Pytorch Hub\footnote{https://pytorch.org/hub/pytorch\_vision\_resnet/}. For each Kinetics video, we sample $T=64$ frames and generate Object-Score following Eqn~\ref{eqn:yobj}.

\paragraph{Hand-object detector $M^{int}$ for Interaction-Map.} We adopt a pre-trained hand-object detector~\cite{shan2020understanding} to discover hand interaction regions. For the detected bounding-box for hands and interactive objects from Kinetics videos, we keep only high-scoring predictions and eliminate bounding boxes with confidence scores less than 0.5.

\YL{Note that all the three pre-trained egocentric models are either easy to access (off-the-shelf models $M^{obj}$ and $M^{int}$) or easy to train ($M^{ego}$). Meanwhile, our auxiliary losses do not require the modification of the network, thus our model can be directly used as a drop-in replacement for downstream egocentric video tasks after pre-training. }

\section{Details: Pre-training on Kinetics}\label{sec:supp_pretrain}
We follow the training recipe in~\cite{slowfast} when training on Kinetics, and use the same strategy for both Slow-only and SlowFast backbones and different implemented methods. 

All the models are trained from scratch for 200 epochs. We adopt a synchronized SGD optimizer and train with 64 GPUs (8 8-GPU machines). The mini-batch size is 8 clips per GPU. The baseline learning rate is set as 0.8 with a cosine schedule for learning rate decaying. We use a scale jitter range of [256, 320] pixels for input training clips. We use momentum of 0.9 and weight decay of $10^{-4}$.

\section{Details: Fine-tuning on Ego-datasets} \label{sec:supp_finetune}
\paragraph{Charades-Ego.} 
During fine-tuning, we train methods using one machine with 8 GPUs on Charades-Ego. The initial base learning rate is set as 0.25 with a cosine schedule for learning rate decaying. We train the models for 60 epochs in total. Following common practice in~\cite{slowfast}, we uniformly sample 10 clips for inference. For each clip, we take 3 crops to cover the spatial dimensions. The final prediction scores are temporally max-pooled. All other settings are the same as those in Kinetics training.  

\paragraph{EPIC-Kitchens.} We use a multi-task model to jointly train verb and noun classification with 8 GPUs on EPIC-Kitchens~\cite{epic}. The models are trained for 30 epochs with the base learning rate as 0.01. We use a step-wise decay of the learning rate by a factor of $10\times$ at epoch 20 and 25. During testing, we uniformly sample 10 clips from each video with 3 spatial crops per clip, and then average their predictions. All other settings are the same as those in Kinetics training.  

For EPIC-Kitchens-100~\cite{epic-100}, we take the same optimization strategy as EPIC-Kitchens~\cite{epic}, except training with 16 GPUs with a 0.02 base learning rate.

\section{Additional results}  \label{sec:supp_results}

\begin{table*}[t]

\centering
\setlength{\tabcolsep}{4.5pt}
	\begin{tabular}{c|l|cc|cc|cc}
		&	& \multicolumn{2}{c|}{verbs} & \multicolumn{2}{c|}{nouns} & \multicolumn{2}{c}{actions}\\
		\cline{2-8}
	\textbf{S1 (seen)}	& Methods	&   top-1 & top-5 & top-1 & top-5 & top-1 & top-5	\\
    \hline
	 \color{darkgray}{{\multirow{2}{*}{w/ audio}}}	& \color{darkgray}{{Epic-Fusion}~\cite{epic-fusion}} & \color{darkgray}{{64.75}} & \color{darkgray}{{{90.70}}} & \color{darkgray}{{46.03}} & \color{darkgray}{{71.34}}  & \color{darkgray}{{34.80}} & \color{darkgray}{{56.65}} \\
 	& \color{darkgray}{{{Epic-Fusion}~\cite{epic-fusion} (Ensemble)}} & \color{darkgray}{\textbf{66.10}} & \color{darkgray}{\textbf{91.28}} & \color{darkgray}{\textbf{47.80}} & \color{darkgray}{\textbf{72.80}} & \color{darkgray}{\textbf{36.66}} & \color{darkgray}{\textbf{58.62}} \\
		\hline
   	\multirow{3}{*}{w/o audio} &    SlowFast~\cite{slowfast} & 64.57 &	89.67&	45.89&	69.50&	34.67&	54.47 \\
    &    Ego-Exo (Single) & 65.97 & 90.32 & 47.99 & 70.72 & 37.09 & 56.32 \\
    &    Ego-Exo (Ensemble) & \textbf{67.84} & \textbf{90.87} & \textbf{49.61} & \textbf{71.77} & \textbf{38.93} & \textbf{58.08} \\
        \hline
        \hline
    \textbf{S2 (unseen)} & & & & & & &  \\
    \hline
    \color{darkgray}{{\multirow{2}{*}{w/ audio}}} &    \color{darkgray}{{Epic-Fusion}~\cite{epic-fusion}} & \color{darkgray}{{52.69}} & \color{darkgray}{{79.93}} & \color{darkgray}{{27.86}} & \color{darkgray}{{53.87}} & \color{darkgray}{{19.06}} & \color{darkgray}{{36.54}} \\
 	& \color{darkgray}{{{Epic-Fusion}~\cite{epic-fusion} (Ensemble)}} & \color{darkgray}{\textbf{54.46}} & \color{darkgray}{\textbf{81.23}} & \color{darkgray}{\textbf{30.39}} & \color{darkgray}{\textbf{55.69}} & \color{darkgray}{\textbf{20.97}} & \color{darkgray}{\textbf{39.40}} \\
    \hline
    \multirow{3}{*}{w/o audio} &    SlowFast~\cite{slowfast} & 53.91 &	80.81 &	30.15 &	55.48 &	21.58 &	37.56 \\
    &    Ego-Exo (Single) & 55.34 & 81.46 & 31.72 & 58.25 & 22.81 & 40.18 \\
     &   Ego-Exo (Ensemble) & \textbf{56.03} & \textbf{81.15} & \textbf{32.54} & \textbf{60.29} & \textbf{23.22} & \textbf{40.97} \\
        \hline
	\end{tabular}
\vspace{-1mm}
\caption{\textbf{Ego-Exo Ensemble results on EPIC-Kitchens (test set).} Our method outperforms all methods in both seen and unseen settings. 
}
\label{tb:epic-55-ensemble}
\end{table*}

\begin{table*}[tb]
	\begin{center}
	\setlength{\tabcolsep}{4.5pt}
	\begin{tabular}{l|cccccc|ccc|ccc}
	    & \multicolumn{6}{c|}{Overall} & \multicolumn{3}{c|}{Unseen Participants} & \multicolumn{3}{c}{Tail Classes} \\
	   \cline{2-13}
	    & \multicolumn{3}{c|}{top-1} & \multicolumn{3}{c|}{top-5} &	\multicolumn{3}{c|}{top-1} & \multicolumn{3}{c}{top-1} \\
	   \cline{2-13}
		Methods &  verbs & noun & \multicolumn{1}{c|}{action} & verb & noun & action & verb & noun & action & verb & noun & action \\
	    \hline
	    Leaderboard1$^\dagger$ & 66.63 & 48.98 & 38.59 & 89.94 & 73.84 & 58.62 & 60.56 & 43.58 & 31.63 & 29.80 & 15.02 & 12.97 \\ 
	    Leaderboard2$^\dagger$ & 65.32 & 47.80 & 37.39 & 89.16 & 73.95 & 57.89 & 59.68 & 42.51 & 30.61 & 30.03 & 16.96 & 13.45 \\
		\hline
         SlowFast~\cite{slowfast} & 63.89 & 49.66 & 37.42 & 88.71 & 74.99 & 58.17 & 57.37 & 44.31 & 29.71 & 33.57 & 22.57 & 16.55 \\
         Ego-Exo (Single) & 66.07 & 51.51 & 39.98 & 89.39 & 76.31 & 60.68 & 59.83 & 45.50 & {32.63} & {33.92} & 22.91 & {16.96} \\
         Ego-Exo (Ensemble) & \textbf{67.13} & \textbf{53.94} & \textbf{42.08} & \textbf{90.07} & \textbf{77.83} & \textbf{62.69} & \textbf{61.05} & \textbf{49.15} & \textbf{35.18} & \textbf{34.73} & \textbf{24.92} & \textbf{18.19} \\
        \hline
	\end{tabular}
	\end{center}
\vspace{-2mm}
\caption{\textbf{Ego-Exo Ensemble results on EPIC-Kitchens-100 action recognition test set.} Leaderboard1$^\dagger$ and Leaderboard2$^\dagger$ are the top two methods on the leaderboard at the time of submission (04/15/2021). Our method is best across all categories.
}
\label{tb:supp-epic-100-results}
\end{table*}

\paragraph{EPIC-Kitchens.}
We report results of an Ensemble of four Ego-Exo models on EPIC-Kitchens in Table~\ref{tb:epic-55-ensemble}. Specifically, the Ensemble model includes Ego-Exo and Ego-Exo* with ResNet-50 and ResNet-101 backbones. As shown in Table~\ref{tb:epic-55-ensemble}, the Ensemble Ego-Exo further improves the performance in all categories, and consistently outperforms the Ensemble model of Epic-Fusion. 

Table~\ref{tb:supp-epic-100-results} shows the Ensemble results on EPIC-Kitchens-100. The Ensemble model of Ego-Exo outperforms the current best model on the leaderboard\footnote{https://competitions.codalab.org/competitions/25923\#results} in all categories at the time of submission, especially on noun and action classes with +5\% and +3.5\% improvements on Overall Top-1 metric. Note that even without Ensemble, our single model already ranks the first on leaderboard and achieves better results than the best leaderboard model in most categories.

\begin{figure*}[t]
\begin{subfigure}{0.5\linewidth}
\centering
\includegraphics[width=1.0\linewidth]{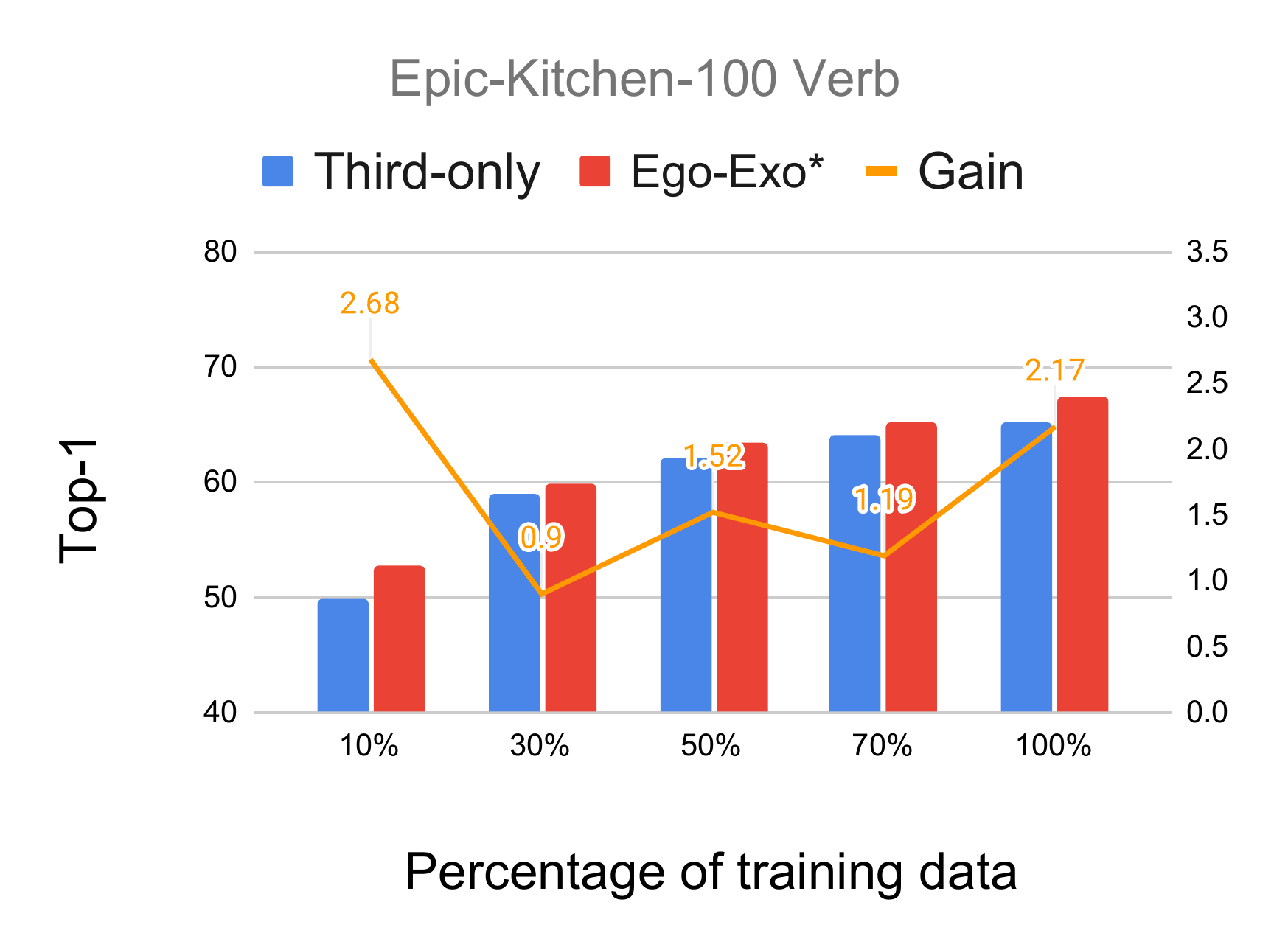}
\end{subfigure}
\begin{subfigure}{0.5\linewidth}
\centering
\includegraphics[width=1.0\linewidth]{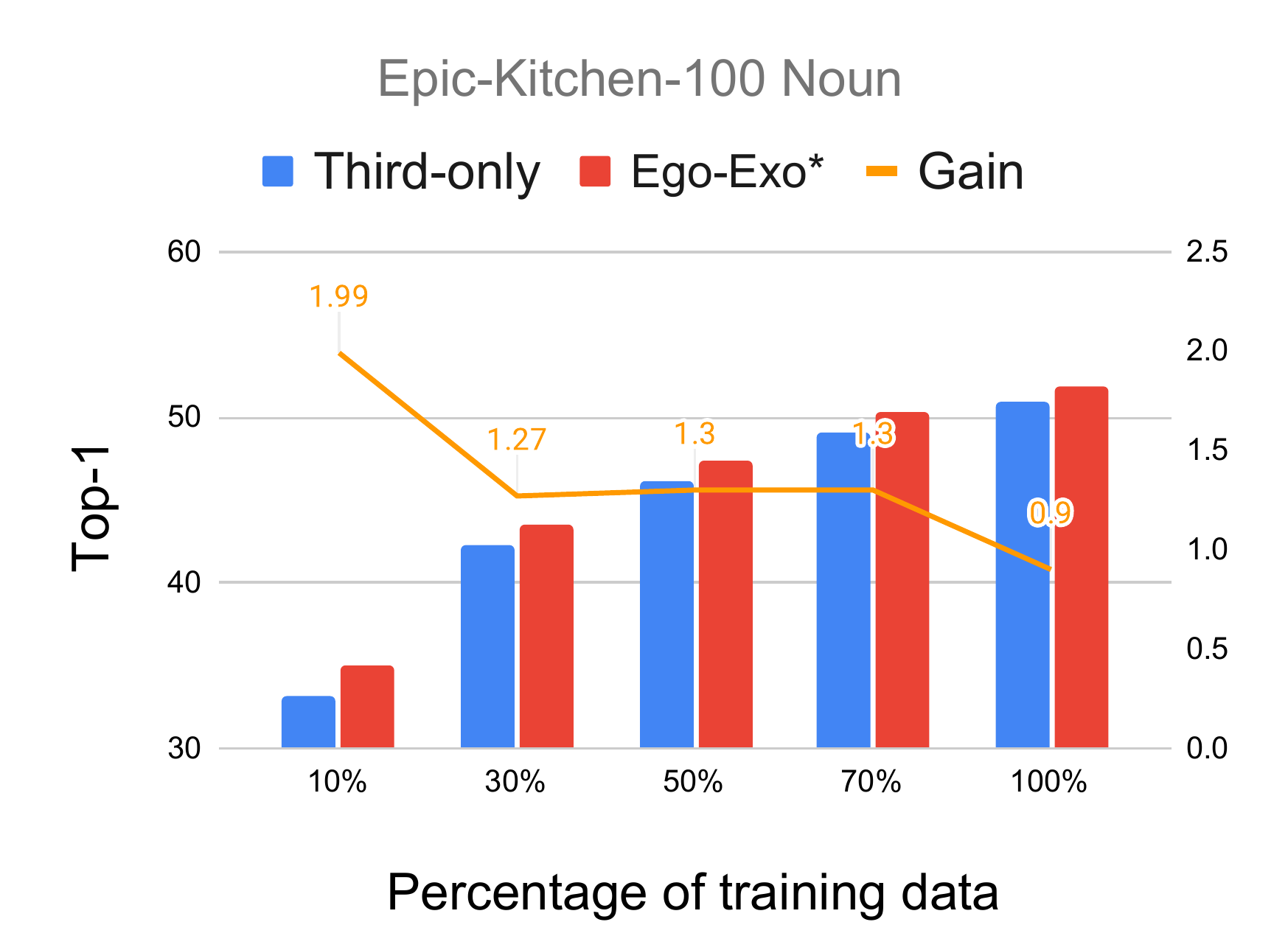}
\end{subfigure}%
\vspace{-2mm}
\caption{\textbf{Performance on EPIC-Kitchens-100 using different percentages of training videos.} Our method consistently improves over the baseline \emph{Third-only} when using different percentages of training videos, with large improvements in very limited data settings (10\% of data). Orange curve shows absolute improvement.
}
\label{fig:partial_results}
\end{figure*}

\YL{
\paragraph{Charades-Ego.} We only train SlowFast and Ego-Exo methods on the egocentric videos from Charades-Ego in Table~\ref{tb:charades-ego-results} of Sec~\ref{sec:exp}. As Charades-Ego also provides third-person videos, we further jointly train first-person and third-person video classification during fine-tuning on Charades-Ego. In this setting, SlowFast and Ego-Exo achieve 25.06 and 28.32 mAP, respectively, with a ResNet-50 backbone. Hence our model further improves over that multi-task setting.
}

\YL{
\paragraph{Something-Something V2 (SSv2).} SSv2~\cite{ssv2} is a non-ego dataset with videos containing human object interactions. We further apply our Ego-Exo method on this dataset and find our \emph{Ego-Exo} method improves over baseline \emph{Third-only} (59.49\% $\rightarrow$ 60.41\% in accuracy). Though our goal remains to address egocentric video, it does seem our method can even have impact beyond it and works for general interaction scenario.
}

\section{Additional Ablation studies} \label{sec:supp_ablation}
\paragraph{Effect of hand-map and object-map in Interaction-Map.} We ablate the Interaction-Map task $M^{int}$ by only using the Hand-Map and Object-Map scores in Eqn~\ref{eq:int-map}. As shown in Table~\ref{tb:interaction-map}, using \emph{Hand-Map} or \emph{Object-Map} alone consistently improves over the baseline (\emph{Third-only}) while combining them (\emph{Interaction-Map}) achieves the best results overall.

\begin{table}[t]
	\begin{center}
	\setlength{\tabcolsep}{4.5pt}
	\begin{tabular}{l|c|cc|cc}
		    &  C-Ego & \multicolumn{2}{c|}{EPIC verbs} & \multicolumn{2}{c}{EPIC nouns} \\
		\cline{2-6}
		Methods	&  mAP  & top-1 & top-5 & top-1 & top-5	\\
		\hline
        Third-only & 24.69 & 61.19 & 87.49 & 46.18 & 69.72 \\
        Hand-Map & 25.28 & 61.35 & 88.02 & 47.33 & \textbf{70.03}	  \\
        Object-Map & \textbf{26.15} & 61.32 & 87.66 & 46.65 & 69.56 \\
        Interaction-Map & 25.91 & \textbf{62.55} & \textbf{88.50} & \textbf{47.71} & 69.62 \\
		\hline				
	\end{tabular}
	\end{center}
\vspace{-4mm}
\caption{\textbf{Ablation study on Interaction-Map task.} 
Combining Hand-Map and Object-map (Interaction-Map) achieves better results overall. Values are averaged over 3 runs. 
}
\vspace{-3mm}
\label{tb:interaction-map}
\end{table}

\begin{table}[t]
\small
	\begin{center}
	\setlength{\tabcolsep}{4pt}
	\begin{tabular}{l|l|c|cc|cc}
	\multicolumn{2}{c|}{}   &  C-Ego & \multicolumn{2}{c|}{EPIC verbs} & \multicolumn{2}{c}{EPIC nouns} \\
		\cline{3-7}
		\multicolumn{2}{c|}{Methods}	&  mAP  & top-1 & top-5 & top-1 & top-5	\\
		\hline
       \multicolumn{2}{c|}{Third-only} & 24.69 & 61.19 & 87.49 & 46.18 & 69.72 \\
        \hline
        \multirow{2}{*}{pre-trained} & $M^{ego}$  & 23.29 & 61.95 & 87.07 & 46.09 & 68.88	  \\
        & $M^{obj}$  & 22.17 & 57.34 & 86.63 & 45.44 & 68.28 \\
        \hline
        \multirow{4}{*}{append} & $M^{ego}$  & 24.92	&60.87	&87.38	&46.40&	69.97		  \\
        &$M^{obj}$  & 24.73	&61.08	&87.35	&45.80&	68.97 \\
        &$M^{int}$ & 24.68	&61.85&	87.39	&46.89&	69.98 \\
        &3 aux & 24.75 & 61.05 & 87.45 & 46.41 & 70.02\\
        \hline
        \multirow{4}{*}{distilled} & $M^{ego}$ & 25.01 & 62.22 & 87.78 & 46.26 & 68.76 \\
        &$M^{obj}$ & 25.49 & 61.65 & 87.57 & 46.27 & 69.52\\
        &$M^{int}$    & 25.91 & 62.55 & \textbf{88.50} & 47.71 & 69.62\\
        &Ego-Exo   & \textbf{26.23} & \textbf{62.83} & 87.63 & \textbf{48.15} & \textbf{70.28}  \\
		\hline				
	\end{tabular}
	\end{center}
\vspace{-3mm}
\caption{\YL{\textbf{Comparison with fine-tuning or appending features from auxiliary task models.} 
Our distillation methods outperform the other two schemes.
}}
\vspace{-3mm}
\label{tb:init_by_aux}
\end{table}

\paragraph{Effect of taking $M^\tau$ as pre-trained model.} In section~\ref{sec:tasks}, we introduce several auxiliary egocentric tasks $M^\tau$ and distill information from them into the video model using auxiliary losses in our Ego-Exo pre-training framework. An alternative way to exploit these signals is to directly use these models ($M^{\tau}$) from auxiliary egocentric tasks $M^\tau$ as our pre-trained models, then fine-tune them on the egocentric datasets. Specifically, we take the ego-classifier $M^{ego}$ and object recognition model $M^{obj}$ as pre-trained models. We do not include hand-object detector $M^{int}$ here as the detection backbone is not compatible with the video backbone.

As shown in Table~\ref{tb:init_by_aux}, though auxiliary task models capture specific egocentric properties, directly use them as pre-trained models is still insufficient. Our methods successfully embed the egocentric information from these auxiliary tasks into the video model through distillation losses, and still enjoy the strong representations learned from the large-scale third-person dataset.

\YL{
\paragraph{Effect of appending embeddings from $M^\tau$.} Another alternative way to exploit these information from auxiliary tasks ($M^\tau$) is to directly use the extracted features on egocentric datasets using $M^{\tau}$. Specifically, we extract the embeddings after the global pooling layer of the three auxiliary models and concatenate them with the ego models during fine-tuning. The results are shown in the `append` rows of Table~\ref{tb:init_by_aux}. It indicates these baselines are less effective than the proposed distillation scheme in our Ego-Exo method.
}

\paragraph{Impact of the scale of egocentric datasets.} We study our model performance under varying scales of egocentric video supervision by using different percentages of videos in EPIC-Kitchens-100~\cite{epic-100}. Fig~\ref{fig:partial_results} shows that our model consistently outperforms the baseline \emph{Third-only} at all dataset scales, though both models perform worse with less egocentric videos during fine-tuning. When using only 10\% of training data, our method improves over the \emph{Third-only} by +2.68\% and 1.99\% on verb and noun tasks.

\YL{\paragraph{Implicit pairing information in Ego-Scores.}
We train the Ego-Classifier $M^{ego}$ for Ego-Scores on Charades-Ego~\cite{charades-ego}. During training, we only use the binary label to indicate the instance is egocentric or not and don't utilize any pairing information. However, Ego-Score might still contains some implicit pairing information. Here, we conduct a ablation study by only taking one view (either ego or non-ego instance) for each pair in Charades-Ego when training $M^{Ego}$. Table~\ref{tb:supp_implicit} shows that methods without any implicit pairing information achieves similar performance. This demonstrates that the implicit pairing information is not critical for our Ego-Exo method.
}

\begin{table}[t]
	\begin{center}
	\setlength{\tabcolsep}{4pt}
	\begin{tabular}{l|c|cc|cc}
		    &  C-Ego & \multicolumn{2}{c|}{EPIC verbs} & \multicolumn{2}{c}{EPIC nouns} \\
		\cline{2-6}
		Methods	&  mAP  & top-1 & top-5 & top-1 & top-5	\\
 		\hline
        Third-only & 24.69 & 61.19 & 87.49 & 46.18 & 69.72 \\
        \hline
        Ego-Score (no-pair) & 24.88 & 62.22 & 87.36 & 46.16 & 68.10 \\
        Ego-Score & 25.01 & 62.22 & 87.78 & 46.26 & 68.76 \\
        \hline
        Ego-Exo (no-pair) & 26.29 & 62.72 & 87.61 & 48.07 & 70.31  \\
        Ego-Exo  & 26.23 & 62.83 & 87.63 & 48.15 & {70.28}  \\
		\hline				
	\end{tabular}
	\end{center}
\vspace{-4mm}
\caption{\textbf{Ablation study on the implicit pairing information}. Methods achieves similar results with or without implicit pairing information. Note that all the methods do not use explicit pairing information from Charades-Ego.
}
\vspace{-2mm}
\label{tb:supp_implicit}
\end{table}

\YL{\paragraph{Ego-Exo with different backbone networks.} Besides using Slow and SlowFast backbones in Sec~\ref{sec:exp}, Table~\ref{tb:supp_backbone} further compares the results using I3D and TSM as the backbone structure. \emph{Ego-Exo} achieves better results over the baseline \emph{Third-only} on both Charades-Ego and Epic-Kitchen datasets. It indicates the versatility of our idea wrt the chosen backbone.
}

\begin{table}[t]
	\begin{center}
	\setlength{\tabcolsep}{4pt}
	\begin{tabular}{l|c|cc|cc}
		    &  C-Ego & \multicolumn{2}{c|}{EPIC verbs} & \multicolumn{2}{c}{EPIC nouns} \\
		\cline{2-6}
		Backbones	&  mAP  & top-1 & top-5 & top-1 & top-5	\\
		\hline				
		Third-only I3D & 25.07 & 59.42 & \textbf{87.82} & 47.16 & 69.39 \\
		Ego-Exo I3D &  \textbf{26.61} &  \textbf{61.51} & 87.28 &\textbf{47.87} & \textbf{69.60}\\
		\hline
		Third-only TSM & 25.66 & 60.17 & \textbf{87.75} & 46.58 & 70.13\\
		Ego-Exo TSM & \textbf{26.22}   & \textbf{61.80} & 87.60 & \textbf{47.71} & \textbf{70.30}\\
		\hline
	\end{tabular}
	\end{center}
\vspace{-2mm}
\caption{\textbf{Results of using I3D and TSM backbones.} \emph{Ego-Exo} consistently outperforms \emph{Third-only}.}
\label{tb:supp_backbone}
\end{table}



\section{Additional qualitative results}  \label{sec:supp_qualitative}

\begin{figure}[t]
\centering
\includegraphics[width=1.0\linewidth]{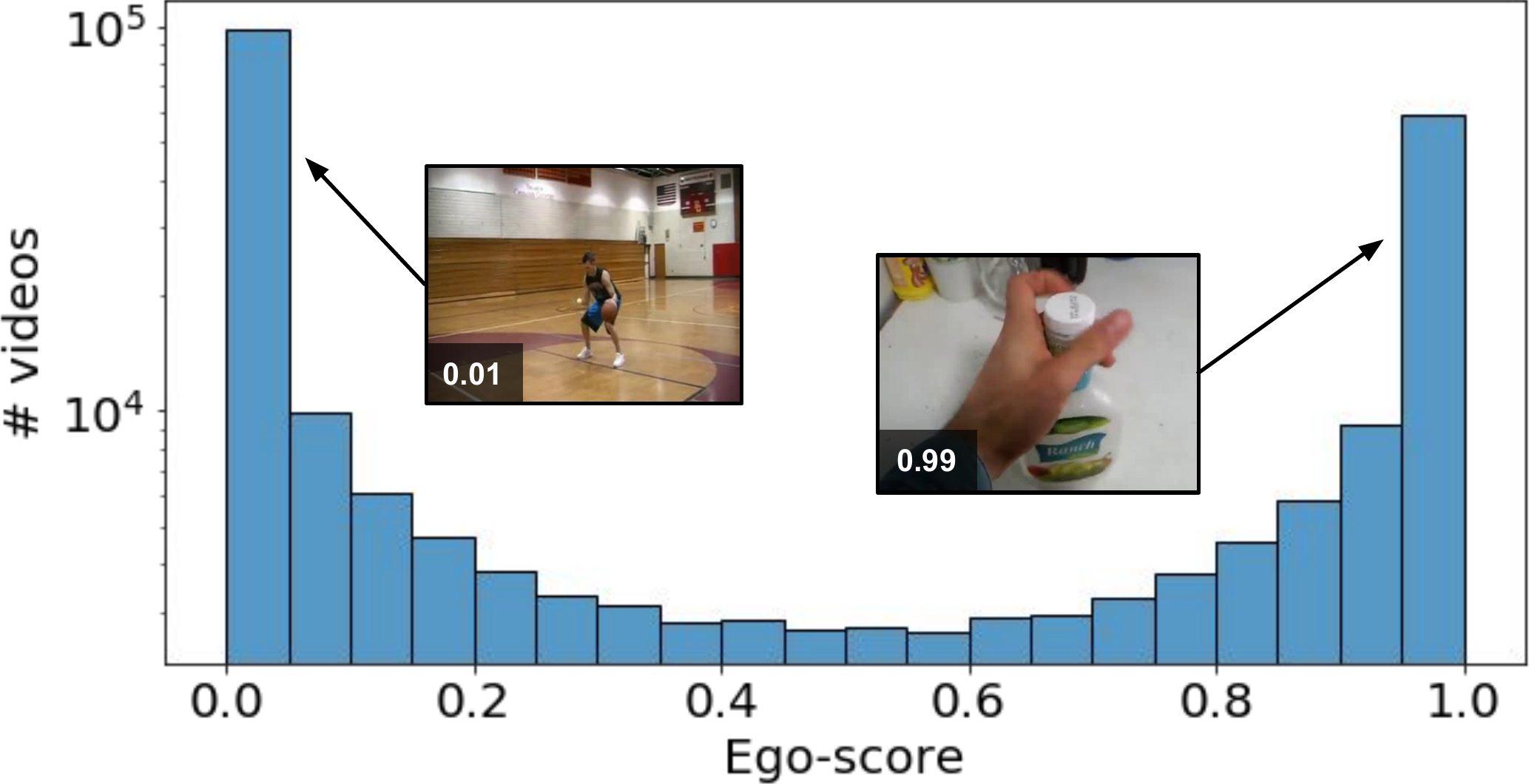}
\caption{\textbf{Distribution of Ego-Scores across Kinetics instances.} Despite being from the third-person perspective, videos in Kinetics display egocentric properties captured by the Ego-Score.}\label{fig:ego_score_dist}
\end{figure}

\paragraph{Distribution of Ego-Score over Kinetics.}
As mentioned in Sec.~\ref{sec:tasks}, though Kinetics videos are predominantly captured in the third-person perspective, the Ego-Score generated by the pretrained classifier $M^{ego}$ is not trivially low for all video instances. Fig~\ref{fig:ego_score_dist} plots the distribution of values this score takes. While a majority of instances have very low scores (not \emph{ego-like}), a large number of instances prominently feature egocentric signals (image inset, right) and have higher scores.

\begin{figure}[t]
\centering
\includegraphics[width=0.9\linewidth]{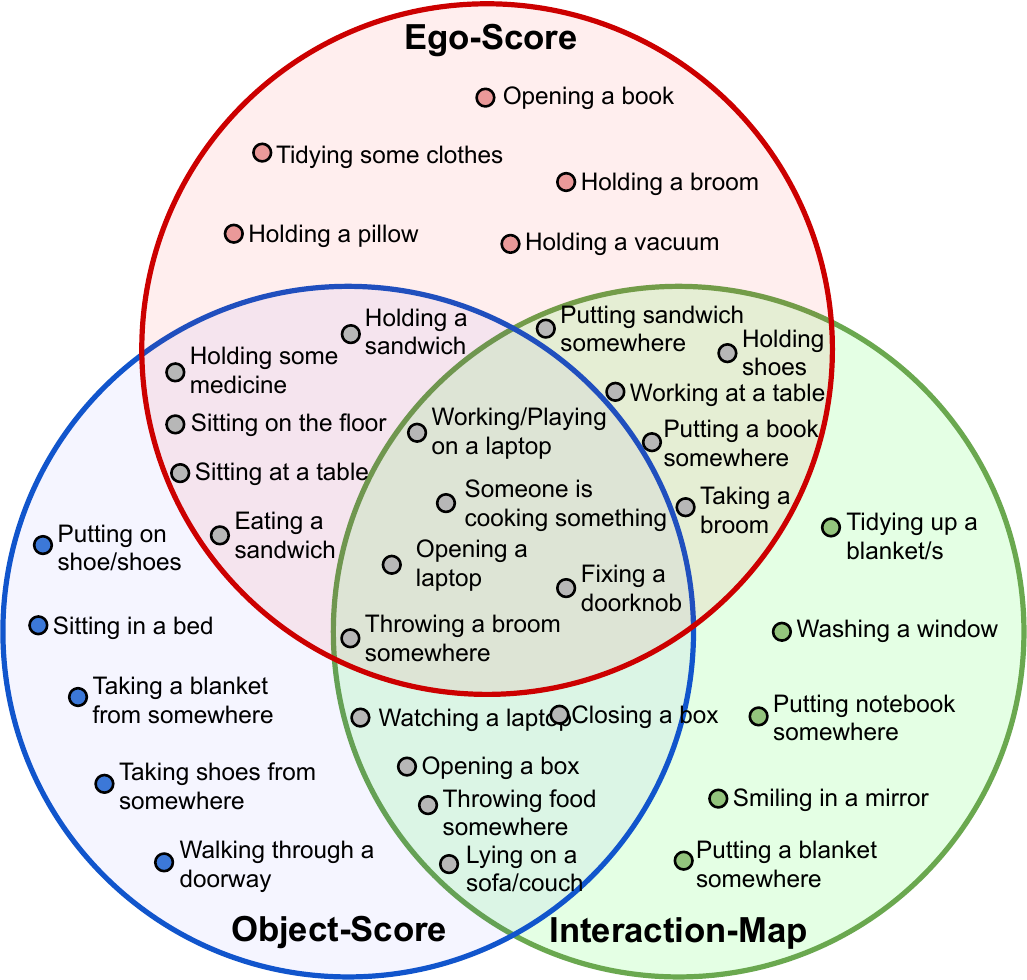}
\caption{\textbf{Charades-Ego classes improved by each egocentric signal.} Each circle contains the classes improved over \emph{Third-only} by a particular ablated model from Table~\ref{tb:different_aux}.}\label{fig:venn_diagram}
\end{figure}

\begin{figure*}[t]
\centering
\includegraphics[width=1.0\linewidth]{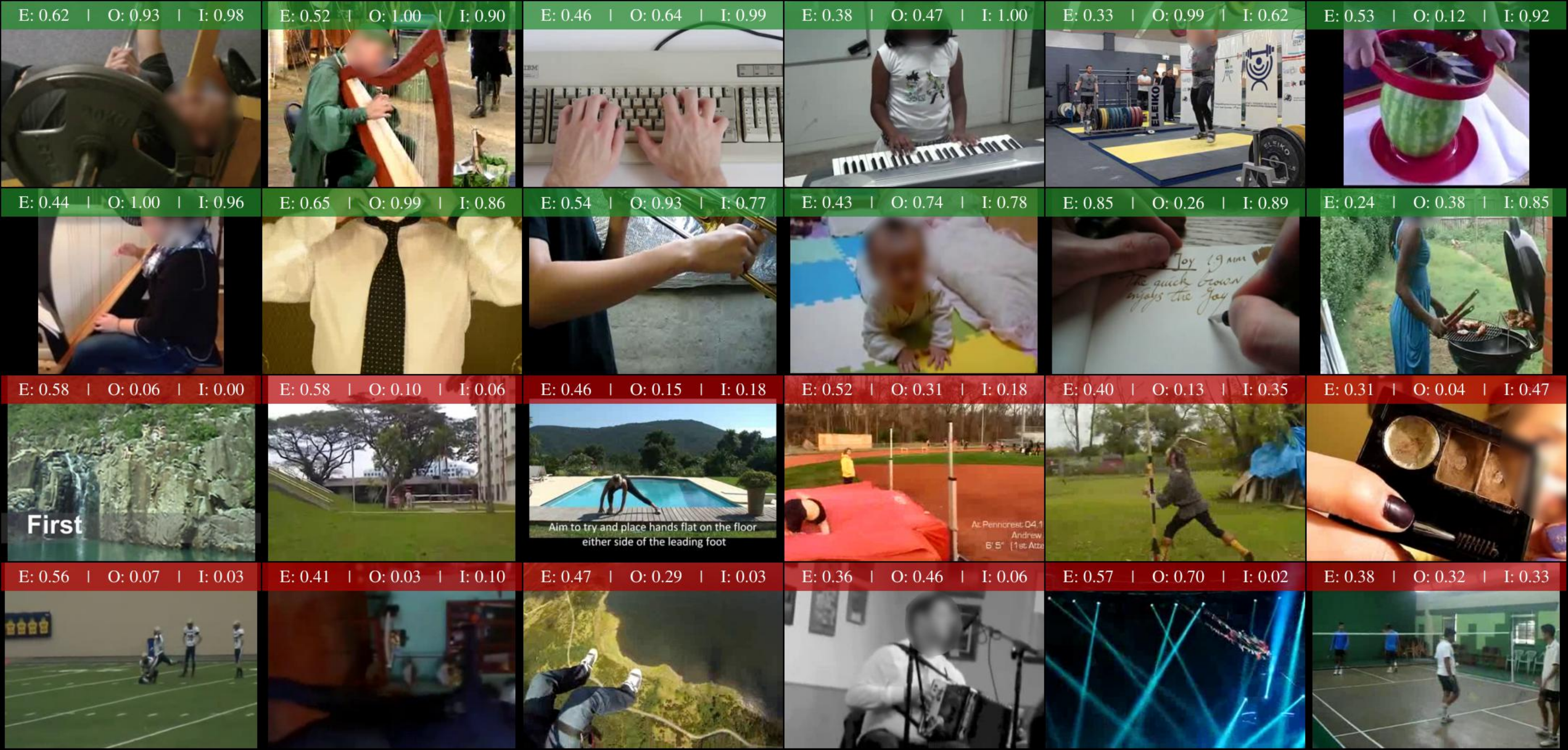}
\caption{\textbf{Additional Kinetics instances sorted by Ego-Exo scores.} Top two rows: Instances that prominently feature hands/objects/egocentric-like motion patterns. Bottom two rows: Instances that feature static scenes devoid of egocentric-like activity.}\label{fig:supp_qual}
\end{figure*}

\paragraph{Class-wise breakdown of improvements from each egocentric task $M^\tau$.}
We present a qualitative result corresponding to the ablation experiment in Table~\ref{tb:different_aux} in the main paper. Fig.~\ref{fig:venn_diagram} shows a venn diagram where each circle contains classes from Charades-Ego that a particular ablated model in Table~\ref{tb:different_aux} improves upon, over the baseline model. For example, the red circle is a model with only Ego-score (row 2, Table~\ref{tb:different_aux}). The overlapping regions between two circles contain classes that are improved by both corresponding ablated models. Note that the three ablated models all contains some classes which are only improved by one particular ablated model, which suggests that three auxiliary tasks capture different egocentric properties.

\paragraph{Additional qualitative examples}
In Fig.~\ref{fig:supp_qual}, we show additional examples of instances from Kinetics, sorted by the scores generated by our pre-trained egocentric models $M^{\tau}$ to supplement Fig.~\ref{fig:retrieval-swatch} in the main paper. The top two rows contain instances with high scores (more ego-like, more prominently features objects, and more hand-object interactions), while the bottom two rows feature instances with low scores. Note that the instances shown are frames from the corresponding video clips. Typically, video clips with more ego-like viewpoint and motions usually have higher Ego-Score. Please see the animated version of this figure in the supplementary video. 





\end{document}